\pdfoutput=1
\documentclass[11pt]{article}

\usepackage{EMNLP2022}
\usepackage{times}
\usepackage{latexsym}
\usepackage{inconsolata}

\usepackage{adjustbox}
\usepackage{amsmath}
\usepackage{amssymb}
\usepackage{graphicx}
\usepackage{tabularx}
\usepackage{algorithm}
\usepackage{multirow}
\usepackage{tabularx}
\usepackage{color, colortbl}
\usepackage{caption}
\usepackage{booktabs}
\usepackage{subcaption}
\usepackage{mwe}
\usepackage{soul}
\usepackage{url}
\usepackage{amsfonts}
\usepackage{bm}
\usepackage{euflag}
\usepackage{tcolorbox}
\usepackage[T1]{fontenc}
\usepackage[utf8]{inputenc}
\definecolor{LightGray}{gray}{0.96}
\definecolor{LightCyan}{rgb}{0.92,0.968,0.968}
\usepackage{microtype}

\usepackage[noend]{algpseudocode}
\usepackage{todonotes}
\usepackage{hyperref}

\makeatletter
\newcommand*\iftodonotes{\if@todonotes@disabled\expandafter\@secondoftwo\else\expandafter\@firstoftwo\fi}
\makeatother

\newcommand{\argmax}{\operatornamewithlimits{argmax}}
\newcolumntype{Y}{>{\centering\arraybackslash}X}

\usepackage{cleveref}
\crefformat{section}{\S#2#1#3}
\crefformat{subsection}{\S#2#1#3}
\crefformat{subsubsection}{\S#2#1#3}
\crefrangeformat{section}{\S\S#3#1#4 to~#5#2#6}
\crefmultiformat{section}{\S\S#2#1#3}{ and~#2#1#3}{, #2#1#3}{ and~#2#1#3}
\Crefformat{figure}{#2Fig.~#1#3}
\Crefmultiformat{figure}{Figs.~#2#1#3}{ and~#2#1#3}{, #2#1#3}{ and~#2#1#3}
\Crefformat{table}{#2Table~#1#3}
\Crefmultiformat{table}{Tabs.~#2#1#3}{ and~#2#1#3}{, #2#1#3}{ and~#2#1#3}
\Crefformat{appendix}{Appx.~\S#2#1#3}
\crefformat{algorithm}{Alg.~#2#1#3}

\newcommand{\sparagraph}[1]{\vspace{0.0mm}\noindent\textbf{#1.}}
\newcommand{\rparagraph}[1]{\vspace{1.8mm}\noindent\textbf{#1.}}
\definecolor{Gray}{gray}{0.92}

\newcommand{\vecmap}{\textsc{VecMap}\xspace}
\newcommand{\blice}{\textsc{blice}\text{r}\xspace}

\usepackage{graphicx,calc}
\newlength\myheight
\newlength\mydepth
\settototalheight\myheight{Xygp}
\newcommand*\inlinegraphics[1]{%
  \settototalheight\myheight{Xygp}%
  \settodepth\mydepth{Xygp}%
  \raisebox{-\mydepth}{\includegraphics[height=\myheight]{#1}}%
}

\title{Improving Bilingual Lexicon Induction with Cross-Encoder Reranking}

\author{Yaoyiran Li, Fangyu Liu, Ivan Vuli\'{c}\thanks{$^*$ Equal senior contribution.}, \textnormal{and} Anna Korhonen\footnotemark[1] \\
  Language Technology Lab, TAL, University of Cambridge \\
  \texttt{\{yl711,fl399,iv250,alk23\}@cam.ac.uk} \\} 
\begin{document}
\maketitle
\begin{abstract}
Bilingual lexicon induction (BLI) with limited bilingual supervision is a crucial yet challenging task in multilingual NLP. Current state-of-the-art BLI methods rely on the induction of cross-lingual word embeddings (CLWEs) to capture cross-lingual word similarities; such CLWEs are obtained \textbf{1)} via traditional static models (e.g., \vecmap), or \textbf{2)} by extracting type-level CLWEs from multilingual pretrained language models (mPLMs), or \textbf{3)} through combining the former two options. In this work, we propose a novel semi-supervised \textit{post-hoc} reranking method termed \textbf{\blice} (\textbf{BLI} with \textbf{C}ross-\textbf{E}ncoder \textbf{R}eranking), applicable to any precalculated CLWE space, which improves their BLI capability. The key idea is to `extract' cross-lingual lexical knowledge from mPLMs, and then combine it with the original CLWEs. This crucial step is done via \textbf{1)} creating a word similarity dataset, comprising positive word pairs (i.e., true translations) and hard negative pairs induced from the original CLWE space, and then \textbf{2)} fine-tuning an mPLM (e.g., mBERT or XLM-R) in a cross-encoder manner to predict the similarity scores. At inference, we \textbf{3)} combine the similarity score from the original CLWE space with the score from the BLI-tuned cross-encoder. \blice establishes new state-of-the-art results on two standard BLI benchmarks spanning a wide spectrum of diverse languages: it substantially outperforms a series of strong baselines across the board. We also validate the robustness of \blice with different CLWEs.

%% and especially so for low-resource languages

%We achieve this In particular, given an input CLWE space and a seed word dictionary, we create a word-pair silver similarity dataset through polarising the positive word pairs and hard negative pairs induced from CLWE. 
%Then, a multilingual PLM is trained in a cross-encoder manner to
%predict the silver similarity score. 

%During inference, a hybrid similarity score of CLWEs and the self-supervised cross-encoder reranker are used.
%polarised training signal, data augmentation, and input template to improve BLI performance. 
%Our method establishes new state-of-the-art (SotA) results on XLING and PanLex-BLI, two standard BLI benchmarks, beating previous SotA by very significant margins. The approach also shows surprising robustness under different CLWE backbones, bringing 5\%-15\% improvement across the board.

\end{abstract}

\section{Introduction and Motivation}
\label{s:introduction}
%% Paragraph 1: what is BLI? Why mapping-based approaches?

Bilingual lexicon induction (BLI) or word translation is one of the core tasks in multilingual NLP \cite[\textit{inter alia}]{Rapp:1995acl,Gaussier:2004acl,shi2021bilingual,li-etal-2022-improving}, with its applications spanning machine translation \cite{Qi:2018naacl,Duan:2020acl}, language acquisition and learning \cite{Yuan:2020emnlp}, as well as supporting NLP tasks in low-resource scenarios \cite{Heyman:2018bmc,Wang:2022acl}, among others. The predominant approach to BLI is based on the induction of a shared \textit{cross-lingual word embedding} (CLWE) semantic space: word translation is then tackled by searching for the nearest neighbour in the other language. Recent BLI work has largely focused on the so-called \textit{mapping-based} or \textit{projection-based} methods \cite{Mikolov:2013arxiv,Ruder:2019survey}. Their prime advantage is strong performance coupled with largely reduced bilingual supervision, typically spanning only 1k-5k seed word pairs \cite{glavas-etal-2019-properly}. This makes them fitting for weakly supervised setups and low-resource languages.

\begin{figure}[!t]
    \centering
    \includegraphics[width=0.999\linewidth]{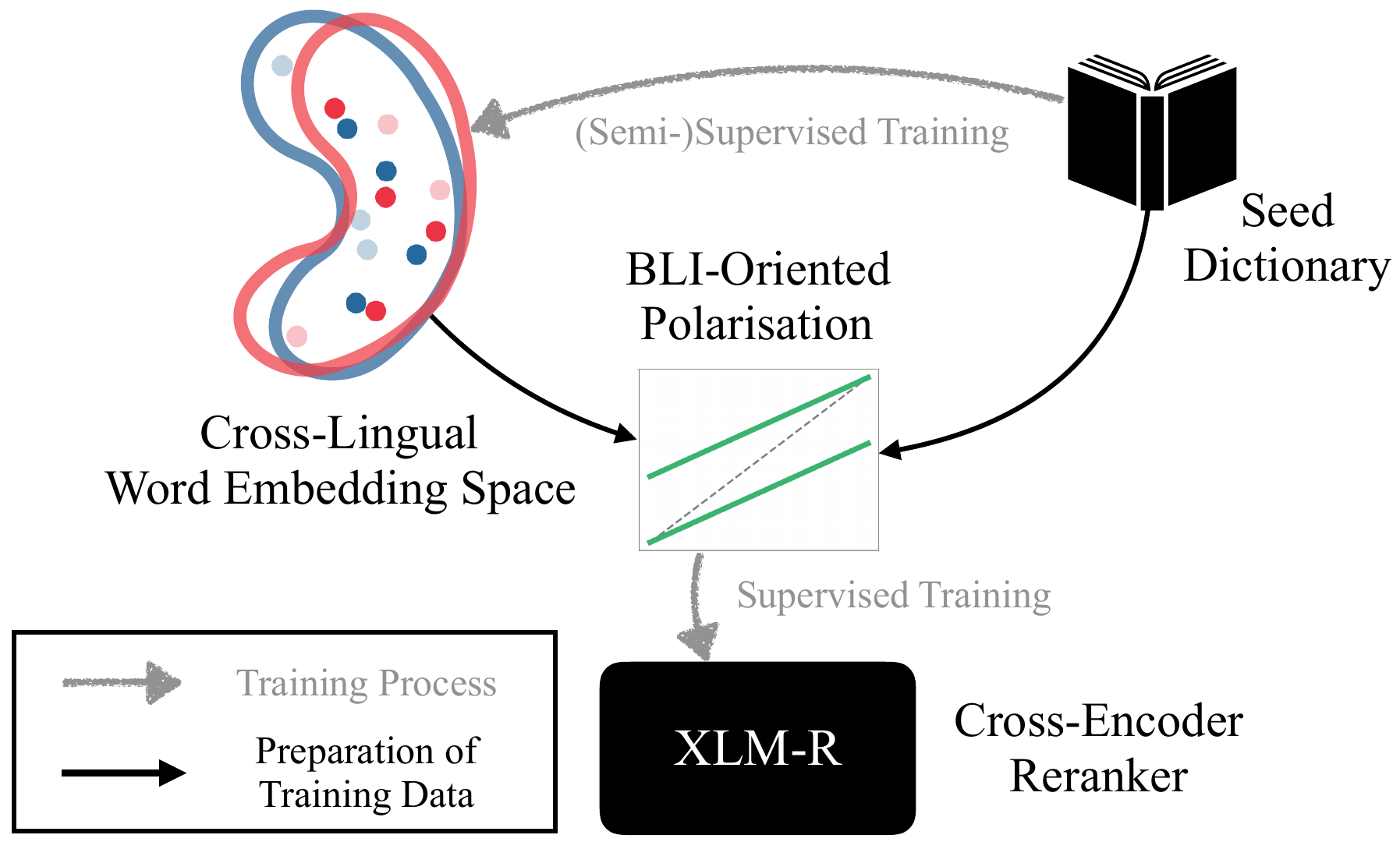}
    %\vspace{-1mm}
    \caption{An overview of the proposed \blice approach, described in detail in \S\ref{s:methodology}.}
    \label{fig:main}
    %\vspace{-2mm}
\end{figure}

%a few thousand seed word pairs, Such methods are particularly suitable for low-resource languages and weakly supervised learning setups: they support BLI with only as much as few thousand word
%translation pairs (e.g., 1k or at most 5k) as the only bilingual supervision \cite{Ruder:2019survey}.\footnote{In the extreme, \textit{fully unsupervised} mapping-based BLI methods can leverage monolingual data only without any bilingual supervision \cite[\textit{inter alia}]{conneau2017word,artetxe2018robust,hoshen2018nonadversarial,Mohiuddin2019revisiting,Ren:2020acl}.

%Research efforts in this area have been focused on \textbf{1)} improving monolingual word embeddings or \textbf{2)} learning more effective methods for aligning monolingual spaces into the CLWE space. representations or producing better alignment between cross-lingual representation spaces. 

%The Earlier works train high-quality monolingual static word embeddings and then map the two languages to a shared cross-lingual space through linear or non-linear mappings trained with a seed dictionary \citep{xing-etal-2015-normalized,lample2019cross,joulin-etal-2018-loss}. With the emergence of multilingual pretrained language models (mPLMs), methods have been introduced for extracting word embeddings from PLMs through either feature extraction \citep{vulic-etal-2020-probing} or self-supervised finetuning \citep{liu-etal-2021-fast}. State-of-the-art BLI is achieved by combining both static word embeddings and PLMs' embeddings for joint inference \citep{li-etal-2022-improving}. 

In parallel, \textit{cross-encoders} (CEs) have gained popularity in sentence-level NLP tasks that involve pairwise sentence comparisons. Unlike the so-called \textit{embedding-based models} (also called bi-encoders or dual-encoders) which process two sequences independently and in parallel to create their embeddings and only model their late interaction \cite{Henderson:2020convert}, CEs take the concatenation of two sequences as input and directly predict the similarity of the two sequences \cite{Humeau2020Poly-encoders:}. As the self-attention heads in CEs can simultaneously attend to tokens from both sequences, CEs are considered more powerful sequence-pair models than embedding-based models which can only perform \textit{post-hoc} comparisons in the embedding space.
A large volume of evidence suggests that, under the same amount of supervision, CEs typically substantially outperform embedding-based models in information retrieval \citep{qu-etal-2021-rocketqa}, dialogue \citep{urbanek-etal-2019-learning}, and semantic similarity tasks \citep{thakur-etal-2021-augmented,liu2022trans}, and their benefits are especially pronounced in low-data regimes with limited task supervision \citep{geigle-etal-2022-retrieve}.

Motivated by the work on CEs in sentence-level tasks, in this work, we propose to use CEs to benefit BLI. In a nutshell, we aim to expose useful word translation knowledge from multilingual pretrained language models (mPLMs) such as mBERT or XLM-R via BLI-oriented CE fine-tuning; this knowledge complements the knowledge stored in the CLWEs. We demonstrate that CEs can be effectively leveraged with cross-lingual word pairs, learning more accurate cross-lingual word similarity scores required for BLI. We present \textbf{\blice} (\textbf{BLI} with \textbf{C}ross-\textbf{E}ncoder \textbf{R}eranking), a \textit{post-hoc} reranking method for BLI, illustrated in Figure~\ref{fig:main} and described in detail in \S\ref{s:methodology}. \blice requires no additional supervision beyond the seed dictionary used for inducing (i.e., mapping) the CLWE space, and it can be combined with any existing CLWE approach, boosting their BLI performance. 

%Fine-tuning multilingual pretrained language models (mPLMs) such as mBERT or XLM-R as BLI-oriented cross-encoders exposes the wealth of word translation knowledge implicitly stored in the mPLMs, which complements the knowledge stored in the CLWEs.

We conduct extensive BLI experiments on two standard BLI benchmarks spanning a diverse language sample, covering 44 translation directions and a total of 352 different BLI setups. We observe large and consistent improvements brought about by \blice across the board: we report gains in 351 out of the 352 BLI setups over the very recent and strong BLI baseline of \newcite{li-etal-2022-improving}, establishing new state-of-the-art (SotA) performance. We also empirically validate that \blice is universally useful, yielding gains with different `CLWE backbones', and run a series of insightful ablations to verify the usefulness of individual components involved in the \blice design. Our code is publicly available at \url{https://github.com/cambridgeltl/BLICEr}.

 \section{Related Work}
\label{s:related_work}
\sparagraph{BLI and CLWEs} 
The predominant BLI methods depend on learning linear or non-linear functions that map monolingual word embeddings to a shared CLWE space \cite{xing-etal-2015-normalized,conneau2017word,joulin-etal-2018-loss,artetxe2018robust,pmlr-v89-grave19a,patra2019bilingual,10.1162/tacl_a_00257,mohiuddin-etal-2020-lnmap,glavas-vulic-2020-non,peng-etal-2021-cross,sachidananda2021filtered}. There have also been attempts to conduct BLI via monolingual and multilingual pretrained language models \cite{gonen-etal-2020-greek,vulic2020multi,vulic-etal-2020-probing,vulic2021lexfit}. However, empirical evidence suggests that these approaches underperform static CLWEs for BLI \cite{vulic-etal-2020-probing}: this is possibly because PLMs are primarily designed for longer sequence-level tasks and thus may naturally have inferior performance in word-level tasks when used off-the-shelf \cite{Vulic:2022arxiv}. Recent work started to combine static and contextualised word representations for BLI \cite{zhang-etal-2021-combining}. In fact, the previous SotA CLWEs for BLI, used as the baseline model in our work, are derived via a two-stage contrastive learning approach combining word representations of both types \cite{li-etal-2022-improving}. Our work builds upon existing CLWE-based BLI methods, and proposes a novel \textit{post-hoc} reranking method that universally enhances BLI performance of any backbone CLWE method.

%\vspace{1.4mm}
\rparagraph{Cross-Encoders} They have wide applications in text matching \cite{chen-etal-2020-dipair},  semantic textual similarity \cite{thakur-etal-2021-augmented,liu2022trans}, and cross-modal retrieval \cite{geigle-etal-2022-retrieve}. They typically outperform the class of Bi-encoder models \cite{Reimers:2019emnlp}, but are much more time-consuming and even prohibitively expensive to run for retrieval tasks directly \cite{geigle-etal-2022-retrieve}. While mPLMs as bi-encoders for BLI have been studied in very recent research \cite{li-etal-2022-improving}, to the best of our knowledge, \blice is the first work to leverage CEs for the BLI task.

\section{Methodology}
\label{s:methodology}
%\subsection{Preliminaries and CLWE Backbones}
%\label{methodology:CLWEs}
\subsection{Background}
\label{methodology:background}
\sparagraph{BLI Task Description} We assume two languages, $L_x$ (source) and $L_y$ (target), with their respective vocabularies $\mathcal{X}$$=$$\{w_{1}^{x},\ldots,w_{|\mathcal{X}|}^{x}\}$ and $\mathcal{Y}$$=$$\{w_{1}^{y},\ldots,w_{|\mathcal{Y}|}^y\}$. Let us denote the set of all possible cross-lingual word pairs as $\bm{\Pi}=\mathcal{X}\times\mathcal{Y}$, where `$\times$' represents the Cartesian product, and a word pair from $\bm{\Pi}$ as $\pi$$=$$(w^{x},w^{y})$. As in a large body of recent BLI work that focuses on mapping-based BLI methods \cite[\textit{inter alia}]{Mikolov:2013arxiv,glavas-etal-2019-properly,li-etal-2022-improving}, we assume (i) $\mathcal{D}_S$, a set of seed word translation pairs for training, and (ii) $\mathcal{D}_T$, a test set of word pairs for evaluation, such that $\mathcal{D}_S,\mathcal{D}_T\subset\bm{\Pi}$ and $\mathcal{D}_S \cap \mathcal{D}_T=\emptyset$. Similar to prior work, we then formulate the BLI task as learning a mapping function $f:\bm{\Pi}\to\mathbb{R}$; $f$ in fact measures cross-lingual word similarity between the words from the input pair $\pi$ \cite{Heyman:2017eacl,Karan:2020acl}. At inference, the BLI task in the $L_x\to 
L_y$ translation direction is then to retrieve the most similar $L_y$ word for each $L_x$ word  $w^{x}$ in $\mathcal{D}_T$: this is the word $\hat{w}^{y}$ from $\mathcal{Y}$ that maximises the cross-lingual similarity score obtained by $f$. More formally:\footnote{It is possible that a single word can have several plausible translations. Following previous work \cite{conneau2017word,glavas-etal-2019-properly,li-etal-2022-improving}, given a query word from the test set $\mathcal{D}_T$, as long as the model retrieves any of the ground-truth translations in its top $K$ predictions, it is considered a correct prediction based on the standard Precision@K (P@K) BLI measure; see also \S\ref{s:experimental}.} $\hat{w}^{y}={\argmax}_{w^{y}\in\mathcal{Y}} \ f(w^{x},w^{y})$. 
%
%% \fixme{YL: this time the task description is wrote in a way that could avoid the one-to-one mapping constraint as possible (our C2 paper got criticism from a reviewer who questions one-to-one mapping).}
%
\iffalse
%\vspace{-1.5mm}
{\footnotesize
\begin{align}
\hat{w}^{y}=\underset{w^{y}\in\mathcal{Y}}{\argmax} \ f(w^{x},w^{y}).
\label{formula:TaskWT}
\end{align}}%
%
\fi
In low-resource setups where only a small seed dictionary $\mathcal{D}_S$ is available as bilingual supervision, most state-of-the-art BLI approaches are still based on the induction of CLWEs \cite{Ruder:2019survey}.

\rparagraph{CLWEs} Let $\bm{X}$$\in$$\mathbb{R}^{|\mathcal{X}|\times d}$ and $\bm{Y}$$\in$$\mathbb{R}^{|\mathcal{Y}|\times d}$ denote the already aligned embedding matrices whose row vectors constitute a shared CLWE semantic space. $L_x$ words are represented by real-valued row vectors/CLWEs of $\bm{X}$, and the representations of $L_y$ words are provided in $\bm{Y}$. In other words, the $d$-dimensional row vector $\mathbf{x}_{i}$ of $\bm{X}$ correspond to the specific word $w_{i}^{x} \in \mathcal{X}$, and the same holds for the target language. 

A plethora of different methods with various data requirements and bilingual supervision can be used to induce such CLWEs \cite{Ruder:2019survey}. Most commonly, due to reduced bilingual supervision requirements, the CLWEs are induced by (i) pretraining monolingual word embeddings independently in two languages, and then (ii) mapping them by linear \cite{Mikolov:2013arxiv,xing-etal-2015-normalized,joulin-etal-2018-loss,artetxe2018robust} or non-linear transformations \cite{glavas-vulic-2020-non,mohiuddin-etal-2020-lnmap}, minimising the distance between the original monolingual word embedding spaces. Optionally, such static CLWEs can be combined or enhanced with external word-level knowledge such as word translation knowledge embedded in multilingual language models \cite{zhang-etal-2021-combining,li-etal-2022-improving,Vulic:2022arxiv}. 

%%Our work take off-the-shelf CLWEs from prior work as the starting point.

%With CLWEs computed by any existing method (e.g., RCSLS, VecMap, ContrastiveBLI)
The actual similarity function $f(\pi)$ is detached from the chosen CLWE method to obtain $\bm{X}$ and $\bm{Y}$. Following prior work \cite{li-etal-2022-improving}, $f(\pi)$ is the Cross-domain Similarity Local Scaling (CSLS) measure \cite{conneau2017word} between the associated embeddings $\mathbf{x}$ and $\mathbf{y}$ which are row vectors of $\bm{X}$ and $\bm{Y}$ respectively:
%
%\vspace{-2.5mm}
%{\footnotesize
\begin{align}
f_C(\pi)=\cos(\mathbf{x},\mathbf{y})-\Gamma_{\bm{X}}(\mathbf{y})-\Gamma_{\bm{Y}}(\mathbf{x}).
\label{formula:CSLSWT}
\end{align}
%}%
%
Here, $cos$ denotes the cosine similarity, $\Gamma_{\bm{X}}(\mathbf{y})$ is the average cosine similarity between $\mathbf{y}$ and its $k$ nearest neighbours (typically $k=10$) in $\bm{X}$; $\Gamma_{\bm{Y}}(\mathbf{x})$ is defined similarly. CSLS is a standard similarity function in BLI which typically outperforms the `vanilla' cosine similarity, as it mitigates the hubness problem during inference.\footnote{We linearly scale $f_{C}$ scores to the range of $[0,1]$.
%For simplicity, we do not elaborate on the formula here, but 
In the rest of this paper, unless stated otherwise, we assume that all $f_{C}$ scores are already scaled.%\cite{conneau2017word}.
}

%% is a highly versatile \textit{post-processing}  method, which can be applied to any CLWE space, that is, on the matrices $\bm{X}$ and $\bm{Y}$ which represent the space, regardless of the chosen method to `precalculate' the CLWE space. 

\subsection{\blice: Cross-Encoder Reranking}
\label{methodology:CER}
\sparagraph{Method in a Nutshell} The proposed \blice method is illustrated in Figure~\ref{fig:main}. The main idea is to refine the initial cross-lingual word similarity scores obtained from the original CLWE space (see Eq.~\ref{formula:CSLSWT}). In particular, assuming the seed dictionary $\mathcal{D}_S$ and the precalculated CLWEs, we first derive positive translation pairs $\mathcal{D}_P\supseteq \mathcal{D}_{S}$ (true translation pairs) and hard negative pairs $\mathcal{D}_{N}$ (semantically similar words that do not constitute a real translation pair). We then \textit{polarise} the scores for both $\mathcal{D}_P$ and $\mathcal{D}_N$ word pairs: the polarisation step effectively increases semantic similarity scores between positives and decreases them for negatives. We then use the polarised scores to fine-tune any mPLM (e.g., mBERT or XLM-R) to transform them into BLI-oriented cross-encoders: that is, we provide word pairs as input, aiming to predict the correct similarity score. Finally, the mPLMs, now transformed into BLI-focused cross-encoders, produce cross-lingual similarity scores for unseen word pairs, which work in synergy with and refine the similarity scores produced by the original cross-lingual word embeddings.

In what follows, we describe the main components of the full \blice post-processing method.

%% As such, our Reranker is compatible with any existing CLWE-based BLI system. 

\rparagraph{Constructing Sets of Positive and Negative Pairs}
The cross-encoder fine-tuning crucially depends on the positive and negative pair sets $\mathcal{D}_P$ and $\mathcal{D}_{N}$. The construction of $\mathcal{D}_P$ starts from the set of gold translation pairs $\mathcal{D}_S$. Prior work demonstrated that additional highly reliable translation pairs can be extracted automatically from the CLWE space \cite{artetxe2018robust,vulic-etal-2019-really}. We thus follow the approach of \citet{li-etal-2022-improving}, and extract additional $N_{aug}$ high-confidence pairs $\mathcal{D}_{aug}$: they are based on the most frequent $N_{freq}$ source and target words in their respective vocabularies, where we conduct both forward and backward BLI for each of the $N_{freq}$ most frequent words in $\mathcal{X}$ and $\mathcal{Y}$, and then retain word pairs with the highest CSLS matching scores. The final augmentation set $\mathcal{D}_{aug}$ is obtained after removing the duplicates and word pairs that contradict the pairs provided by $\mathcal{D}_S$. The final set of positives is then $\mathcal{D}_P=\mathcal{D}_{S}\cup\mathcal{D}_{aug}$. 

%%In the Eq. \ref{formula:CSLSWT}), and finally remove pairs that contradicts $\mathcal{D}_S$. $\mathcal{D}_P=\mathcal{D}_{S}\cup\mathcal{D}_{aug}$. 

%We target CEs at the `soft spot' of CLWEs, embodied in word pairs with similar or even higher CSLS matching scores than the corresponding positive pairs; we call these hard negative pairs. 

In our preliminary analyses of CSLS similarity scores within the original CLWE space, we have detected that some non-translation pairs actually produce similar or even higher absolute CSLS scores than the corresponding ground truth positive pairs. Providing such information of \textit{hard negative pairs}, collected into the set of negatives $\mathcal{D}_{N}$, would be a strong signal for CE fine-tuning: the core idea is that the cross-encoder will be able to `overturn' such wrong predictions from the original CLWE space. In practice, for each $(w^{x}_{+},w^{y}_{+})\in {D}_P$, we propose to retrieve their respective negative words $w^{x}_{-},w^{y}_{-}$ that satisfy the following:
%
%Through fine-tuning, we hope to turn CEs, known for strong model capabilities, into BLI rerankers, enabling them to produce adjusted and refined matching scores such that previously `intractable' hard negative pairs (for pure CLWE-based methods) will not be misclassified as a translation pairs. 
%
%
%
%\vspace{-1.5mm}
%{\footnotesize
\begin{align}
\left\{\begin{matrix}
f_{C}(w^{x}_{+},w^{y}_{-})\geqslant f_{C}(w^{x}_{+},w^{y}_{+}) - \delta,\\ 
f_{C}(w^{x}_{-},w^{y}_{+})\geqslant f_{C}(w^{x}_{+},w^{y}_{+})- \delta,
\end{matrix}\right.
\label{formula:NegPairs}
\end{align}
%}%
%
where $\delta$ is a tunable margin. For a positive pair, we include at most $N_{neg}$ $w^{x}_{-}$ and $w^{y}_{-}$ words to build negative pairs $(w^{x}_{+},w^{y}_{-}),(w^{x}_{-},w^{y}_{+})\in {\mathcal{D}}_N$. We exclude pairs that already exist in ${\mathcal{D}}_P$, which occasionally occurs due to polysemy; that is, it holds ${\mathcal{D}}_P\cap {\mathcal{D}}_N=\emptyset$.  

We also define a \texttt{reverse} operation $(\cdot)^{\star}$:
$\pi^{\star}=(w^{x},w^{y})^{\star}=(w^{y},w^{x})$. Similarly, we extend the definition on sets such that the \texttt{reverse} of a set is a set of all its elements reversed, e.g., $\bm{\Pi}^{\star}=\mathcal{Y}\times\mathcal{X}$. It is evident that when using a set of fixed CLWEs, it holds $f_{C}(\pi)=f_{C}(\pi^{\star})$ (Eq.~\ref{formula:CSLSWT}). However, CEs are sensitive to the order of languages: for symmetry, we thus also provide $\mathcal{\mathcal{D}}_P^{\star}$ and $\mathcal{\mathcal{D}}_{N}^{\star}$ for CE fine-tuning. Due to the imbalance between $\mathcal{\mathcal{D}}_P$ and $\mathcal{\mathcal{D}}_{N}$, we repeat each positive pair (in both $\mathcal{\mathcal{D}}_P$ and $\mathcal{\mathcal{D}}_P^{\star}$) ${N}_{rep}$ times for CE fine-tuning. The choice of the value ${N}_{rep}$ impacts the distribution of word pairs for CE fine-tuning: $\pi\sim p_{\pi}$.

\rparagraph{CE Fine-Tuning with Polarised Similarity Scores} 
As mentioned, we observed that in the original CLWE space there are frequent cases where a true translation pair \textit{from the training set} obtains a lower CSLS score than one or more hard negative pairs (i.e., non-translations): $f_{C}(w^{x}_{+},w^{y}_{+})<f_{C}(w^{x}_{+},w^{y}_{-})$. Such cases can be fixed or mitigated by \textit{polarising} similarity scores for the word pairs in the sets $\mathcal{D}_P$ and $\mathcal{D}_N$ (and also $\mathcal{D}_P^{\star}$ and $\mathcal{D}_N^{\star}$) before CE fine-tuning. The polarisation step in practice means \textbf{1)} increasing the score $f_{C}(w^{x}_{+},w^{y}_{+})$ of positive pairs, and \textbf{2)} decreasing $f_{C}(w^{x}_{+},w^{y}_{-})$. The key rationale behind polarisation is exactly the following: cross-encoders can resolve or mitigate the difficult cases of highly similar hard negatives, where polarisation enables us to provide the correct learning signal for this purpose. 

%% \textbf{1)} cross-encoders should still (learn to) predict similarity scores in the continuous space (rather than doing binary predictions) so that they can be combined with the original CLWE similarity scores; 

\begin{figure}[t!]
    \centering
    \includegraphics[width=0.88\linewidth]{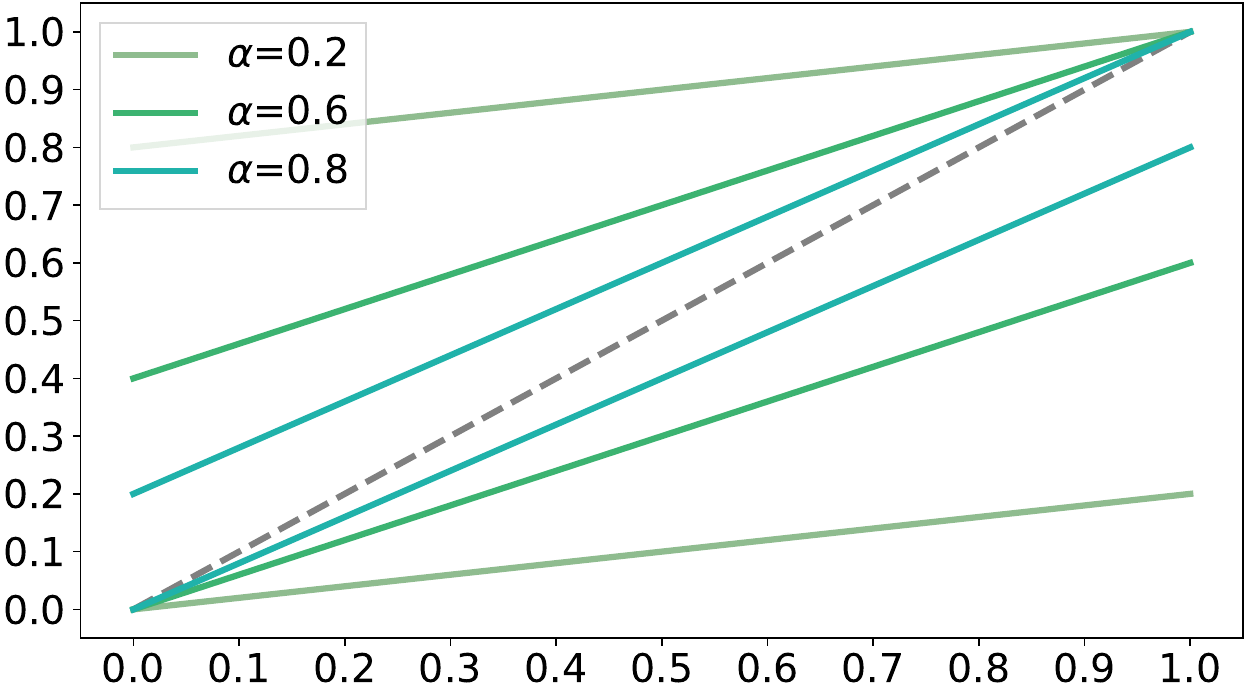}
    %\vspace{-1mm}
    \caption{Linear polarisation functions with different values of the hyper-parameter $\alpha$.}
    \label{fig:polarise}
   % \vspace{-2mm}
\end{figure}

In practice, we construct a pair of monotonically increasing polarisation functions $g^{+}, g^{-}:\mathbb{R}\to\mathbb{R}$ to adjust the CSLS scores from the original CLWE space, such that for a positive word pair it holds $g^{+} \circ f_{C}(\pi) \geqslant f_{C}(\pi)$; for a negative word pair it holds $g^{-} \circ f_{C}(\pi) \leqslant f_{C}(\pi)$. %\footnote{Their domains and codomains are within $[0,1]$ since the output values of CEs are conventionally in the range.} 
%
%%Through the polarised distillation, CEs, known for strong fitting capabilities, can gradually learn to identify real word pairs among hard distractors. %
%
We adopt a pair of simple linear functions, where their domains and codomains are within the $[0,1]$ interval, which is the same interval of the output values of CEs.\footnote{This is because the sigmoid function is applied on the output logits of CEs.} The functions are as follows:
%
%\vspace{-1.5mm}
%{\footnotesize
\begin{align}
\left\{\begin{matrix} 
g^{+}(z)=\alpha\cdot z - \alpha + 1,\\
g^{-}(z)=\alpha\cdot z.\;\;\;\;\;\;\;\;\;\;\;\;\;\;
\end{matrix}\right.
\label{formula:Polarisation}
\end{align}
%}%
%
$\alpha \in [0,1]$ is a hyper-parameter, with its impact on the polarisation function illustrated in Figure~\ref{fig:polarise}.When $\alpha=1$, the original unaltered CSLS  scores are used to fine-tune multilingual LMs; when $\alpha=0$, binary labels ($1$: true translation pairs , $0$: negative pairs) are used for CE fine-tuning. The linear polarisation function can be seen as a more expressive function that generalises over the two special cases.\footnote{In our preliminary experiments, we also investigated non-linear polynomials as polarisation functions, but such functions led to only small to negligible BLI performance gains; we thus omit them for brevity.}

%%generalisation as well as a refinement of the two special cases ($\alpha\in \{0,1\}$).

We then use the adjusted (i.e., polarised) scores to fine-tune multilingual pretrained LMs. For a word pair $\pi$, we denote the CE prediction as $u_{\pi}=\sigma(f_{\theta}(\pi))$, where $\sigma(\cdot)$ is the sigmoid function. The polarised similarity score for fine-tuning is denoted as $v_{\pi}=g^{+/-}\circ f_{C}(\pi)$, where $g^{+/-}(\pi)$ is $g^{+}(\pi)$ if $\pi\in \mathcal{D}_{P}\cup \mathcal{D}_{P}^{\star}$, and $g^{-}(\pi)$ if $\pi\in \mathcal{D}_{N}\cup \mathcal{D}_{N}^{\star}$. $\theta$ represents the entire set of parameters of the cross-encoder. We then train \blice with the standard Binary Cross-Entropy loss:
%
%\vspace{-1.5mm}
%{\footnotesize
\begin{align}
\centering
\mathcal{L}\!=\!&-\mathbb{E}_{\pi\sim p_{\pi}}\![v_{\pi}\!\cdot\!\log(\sigma(u_{\pi})\!)\! \notag \\
&+\!(1\!-\!v_{\pi})\!\cdot\! \log(\!1\!-\!\sigma(u_{\pi})\!)]. \notag
\label{formula:Loss}
\end{align}
%}%
%
%% and we embed $\pi$ in a hand-crafted template as input, which we will introduce in the next paragraph

\rparagraph{Providing Text Input to Cross-Encoders} 
Cross-encoders must `consume' word pairs $\pi$ as input. We enclose the word pairs into text templates that are fed to the CE. Given a word pair $(w^x, w^y)$, the final input is then in the format $T(w^{x}), T(w^{y})$, where a template-based transformation is applied to each word from the pair. There is a spectrum of possible templates (e.g., see Appendix~\ref{app:templates} for a list of 16 templates we designed), and in our main experiments we rely on the following one: $T(w) = [w]\ ([L_w])!$, where $[w]$ is replaced by the actual word, and $[L_w]$ is the word for the language to which $w$ belongs (i.e., more precisely $w$ can be found in the corresponding vocabulary of language $L_w$) in that actual language. For instance, the CE input for an English-French word pair \textit{(apple, pomme)} is \textit{apple (english)!, pomme (français)!}.

In practice, each mPLM's dedicated tokeniser then splits the template text input of two words into two sequences of WordPiece/subword tokens, inserts the special separation token between them, and appends and prepends other special tokens.\footnote{Different CEs may have different input text formats. XLM-R: \texttt{[s] ... [/s] [/s] ... [/s]}; mBERT: \texttt{[CLS] ... [SEP] ... [SEP]}. In the two examples \texttt{...} denotes a subword sequence, \texttt{[s]} and \texttt{[CLS]} are special prefix tokens, and \texttt{[/s]} and \texttt{[SEP]} are special separation/suffix tokens.} 

%%We experiment with $16$ different templates (details in Appendix) and pick the template "[WORD] ([LANG])!" where "[WORD]" is a raw word and "[LANG]" indicates the language. E.g., the templated version of the English-French word pair (apple, pomme) becomes "apple (english)!" and "pomme (français)!".

\rparagraph{Combining Similarity Scores}
At BLI inference, we combine the similarity scores computed by the fine-tuned CE with the original CLWE scores, via a standard linear interpolation:
%Finally, we adopt a CE-enhenced cross-lingual word similarity measure, which is a hybrid of CLWE and CE predictions ($\lambda\in[0,1]$ is a tunable hyper-parameter):
%\footnote{Note that the output value of CEs is practically between $0$ and $1$;see \url{www.sbert.net/examples/applications/cross-encoder/README.html}. Therefore, }
%\vspace{-1.5mm}
%{\footnotesize
\begin{align}
\widetilde{f}_{\theta}(\pi)&= \frac{\sigma(f_{\theta}(\pi)) + \sigma(f_{\theta}(\pi^{\star}))}{2}.\\
f_{\text{Mix}}(\pi)&= (1-\lambda)f_{C}(\pi)+ \lambda\widetilde{f}_{\theta}(\pi),
\label{formula:CEevalMix}
\end{align}
%}%
where $\lambda\in[0,1]$ is a tunable hyper-parameter.

\rparagraph{CE Reranking}
One major drawback of cross-encoders is their high computation overhead for retrieval tasks with a large number of items in the target set \cite[\textit{inter alia}]{karpukhin-etal-2020-dense,Humeau2020Poly-encoders:,geigle-etal-2022-retrieve}. In particular, given a source word $w^{x}$, in order to predict its translation in $\mathcal{Y}$, CE models need to calculate similarity scores over all candidate pairs in $\mathcal{Y}$. We thus follow the body of work in information retrieval \cite{Lin:2021book}, and adopt a more efficient, two-stage `\textit{retrieve-and-rerank}' approach \cite{Karan:2020acl,geigle-etal-2022-retrieve}. First, we use more efficient CLWEs to retrieve the $N_{cand}\ll|\mathcal{Y}|$ candidate pairs with the highest similarity scores, and then rerank them relying on the additional knowledge from the CEs, based on Eq.~\ref{formula:CEevalMix}. Feeding only the $N_{cand}$ pairs to the CEs substantially decreases the computational overhead.

%\subsection{A Universal Multiway BLI Reranker}

\section{Experimental Setup}
\label{s:experimental}
\subsection{BLI Setups and Datasets}
%\sparagraph{BLI Setups and Datasets}
We use two standard and established BLI datasets: \textbf{1)} XLING \cite{glavas-etal-2019-properly} and \textbf{2)} PanLex-BLI \cite{vulic-etal-2019-really}. XLING provides BLI training and test lexicons covering $8$ languages from diverse language families (Croatian: \textsc{hr}, English: \textsc{en}, Finnish: \textsc{fi}, French: \textsc{fr}, German: \textsc{de}, Italian: \textsc{it}, Russian: \textsc{ru}, Turkish: \textsc{tr}). We consider all $14$ EN$\to$$*$ and $*$$\to$EN BLI directions. Further, for each BLI direction, we run experiments in \textit{supervised} settings (i.e., where the training set $\mathcal{D}_S$ covers 5k word pairs) and \textit{semi-supervised} settings (i.e., $|\mathcal{D}_S|$ = 1k).  PanLex-BLI is a BLI benchmark oriented towards low-resource languages. We use a subset of PanLex-BLI comprising six diverse languages (Bulgarian: \textsc{bg}, Catalan: \textsc{ca}, Estonian: \textsc{et}, Georgian: \textsc{ka}, Hebrew: \textsc{he}, Hungarian: \textsc{hu}), yielding a total of $30$ BLI directions. Since we deal with lower-resource languages in PanLex-BLI, we consider only semi-supervised setups with 1k pairs for training. Both XLING and PanLex-BLI trim the vocabularies of each language to the most frequent 200k words, and the standard choice of pretrained fastText word embeddings \cite{bojanowski2017enriching} is used to derive CLWEs (see also Appendix~\ref{app:reproducibility}).   

%To simulate the scarcity of bilingual data between 

Both datasets provide a test set of 2k work pairs for each language pair, without any overlap with the training pairs. We report the standard \textit{Precision@1} (P@1) scores.\footnote{We observed very similar performance trends for P@5 and Mean Reciprocal Rank (MRR) as BLI measures.} CSLS with $k$=$10$ (see Eq.~\ref{formula:CSLSWT}) is used as the CLWE-based similarity function. 

%For CLWE baselines, we derive P@1 scores with CSLS adjustment as described in \Cref{methodology:CLWEs}, on top of which our proposed \blice is applied.

\subsection{CLWE Models (Baselines)}
%\rparagraph{CLWE Models (Baselines)}
Our BLI method is evaluated against a representative set of strong SotA BLI models from recent literature; all of them are CLWE-based and with publicly available implementations.  Here, we provide brief summaries:\footnote{For further technical details and descriptions of each BLI model, we refer to their respective publications. We used the publicly available implementations of all the baseline models.} 

%RCSLS and \vecmap were two well-known SotA BLI methods before ContrastiveBLI \cite{li-etal-2022-improving}. ContrastiveBLI is currently the strongest BLI model outperforming all existing methods in both supervised and semi-supervised BLI: it consists of C1 (pure fastText-based) and C2 (C1+mBERT) variants. We use the same set of fastText WEs provided by XLING and PanLex-BLI for deriving CLWEs. C2 additionally uses mBERT to improve C1's results. $k$=$10$ is used when calculating CSLS matching scores for all CLWE baselines. For all the baselines, we follow their original suggested settings and hyperparameter choices for supervised and semi-supervised BLI setups respectively. 

%We verify (near-)optimal BLI results are derived with the suggested settings. 
%In the following we briefly introduce the four CLWE models that are involved in our comparison.

%\vspace{0.5mm}
\noindent \textbf{RCSLS} \cite{joulin-etal-2018-loss} is a representative CLWE model trained directly with the BLI-style objective and displays strong performance in supervised BLI tasks. It first learns an initial mapping via Procrustes \cite{xing-etal-2015-normalized} and then fine-tunes the mapping via a relaxed CSLS loss.  

%\vspace{0.5mm}
\noindent \textbf{\vecmap} \cite{artetxe2018robust} leverages a self-learning procedure and demonstrates strong performance in unsupervised and semi-supervised BLI settings. It also supports unsupervised BLI. For the supervised setting, its self-learning is switched off, producing better BLI results.

%\vspace{0.5mm}
\noindent \textbf{ContrastiveBLI} \cite{li-etal-2022-improving} is the current SotA BLI model outperforming all existing methods in supervised and semi-supervised BLI settings. It is a two-stage model where both stages, termed \textbf{C1} and \textbf{C2}, leverage contrastive fine-tuning. 
Stage C1 is based purely on static CLWEs (i.e., CLWEs derived from static WEs such as fastText) and it refines an initial CLWE mapping via BLI-oriented contrastive fine-tuning with a self-learning procedure, attracting true translation pairs together and pushing away hard negative pairs. Stage C2 first derives `decontextualised' mBERT word embeddings via contrastively tuning a pretrained mBERT model, and then linearly combines C1-induced CLWEs with mBERT-based word vectors. C2 typically further improves BLI performance over C1's output. In this work, we provide comparisons against CLWEs from both stages.\footnote{\citet{li-etal-2022-improving} empirically validated that other standard BLI methods such as LNMap \cite{mohiuddin-etal-2020-lnmap} or FIPP \cite{sachidananda2021filtered} yield BLI performances which are on average similar to or weaker than those obtained by RCSLS and \vecmap. Moreover, all the methods: RCSLS, \vecmap, FIPP, and LNMap, are consistently outperformed by ContrastiveBLI's C1 and C2 stages \cite{li-etal-2022-improving}, which serve as our strongest BLI baselines. Because of that and for clarity, we omit FIPP and LNMap from the experiments.}

For all the baselines, we follow their original suggested settings and hyper-parameter choices for supervised and semi-supervised BLI settings. 

\subsection{\blice: Training Setup and Hyper-parameters}
%, while we conduct a comparative study on the impact of the chosen mPLM later in \S\ref{s:results}.

%\rparagraph{Unsupervised $\&$ Zero-shot Setups} Our study mainly focuses on supervised and semi-supervised settings which are more practical: \newcite{vulic-etal-2019-really} find that using even a small amount of supervision always outperforms unsupervised methods. But in \Cref{s:unsup} and Appendix \ref{appendix:unsupervised}, we also conduct preliminary investigations on the settings where no bilingual supervision between source and target languages are available. 

\sparagraph{Training Setup} Since BLI datasets typically do not provide separate development sets, previous work conducted hyper-parameter search on a randomly selected language pair \cite{glavas-etal-2019-properly,Karan:2020acl,li-etal-2022-improving} from the BLI benchmark. We adopt this approach, and tune \blice's hyper-parameters on the (\textsc{en},\textsc{tr}) pair from XLING. For PanLex-BLI, we inherit all hyper-parameter values from the XLING experiments, and only further tune the $\lambda$ value on the randomly sampled (\textsc{hu},\textsc{ka}) pair. We also select the final text template (see \S\ref{s:methodology} and Appendix~\ref{app:templates}) in the same fashion.

\rparagraph{Multilingual Pretrained Language Models} We test three mPLMs: mBERT \cite{devlin2018bert}, XLM-R$_{base}$, and XLM-R$_{large}$ \cite{conneau-etal-2020-unsupervised}. Unless noted otherwise, XLM-R$_{large}$ is used as the main model for \blice.

%for better domain adaptation. 
\rparagraph{Hyper-parameters} For supervised setups, \blice is fine-tuned for 3 epochs, $N_{aug}$=$0$, $N_{rep}$=$8$, $\delta$=$0.1$, and $\alpha$=$0.7$; for semi-supervised setups, \blice is trained for $5$ epochs, $N_{aug}$=$4$k, $N_{rep}$=$4$, $\delta$=$0.2$, and $\alpha$=$1.0$. The $\lambda$ values for different setups are listed in Appendix~\ref{appendix:lambda}. In all BLI setups, we use AdamW \cite{Loschilov:2018iclr} with the learning rate of $1.2e$-$5$, and the weight decay is $0.01$; the maximum sequence length is $20$, the batch size is $256$, $N_{freq}$=$20$k, $N_{neg}$=$28$, and $N_{cand}$=$28$ (see \S\ref{s:methodology} again for the description of each hyper-parameter).

\section{Results and Discussion}
\label{s:results}
%\subsection{Main Results}
%\label{results:main}
The main results on XLING and PanLex-BLI are presented in Table~\ref{table:main-xling} and Table~\ref{table:panlex-main}, with additional results available in Appendix~\ref{appendix:panlex}). The tables span $14+30=44$ BLI directions, in supervised and semi-supervised scenarios, and with four CLWE methods respectively, which yields a total of $352$ different BLI setups. One major quantitative finding is that the proposed \blice method derives gains in $351/352$ setups. In what follows, we delve deeper into the analyses across multiple aspects. %discuss our findings from the following aspects.

\begin{table*}[ht]
\def\arraystretch{0.99}
\begin{center}
\resizebox{0.999\textwidth}{!}{%
\begin{tabular}{llllllllllllllll}
\toprule 

\rowcolor{Gray}
\multicolumn{1}{c}{[5k] \bf Pairs}  &\multicolumn{1}{c}{\bf EN$\to$DE}  &\multicolumn{1}{c}{\bf DE$\to$EN} &\multicolumn{1}{c}{\bf EN$\to$FI} &\multicolumn{1}{c}{\bf FI$\to$EN}
&\multicolumn{1}{c}{\bf EN$\to$FR}
&\multicolumn{1}{c}{\bf FR$\to$EN}
&\multicolumn{1}{c}{\bf EN$\to$HR}
&\multicolumn{1}{c}{\bf HR$\to$EN}
&\multicolumn{1}{c}{\bf EN$\to$IT}
&\multicolumn{1}{c}{\bf IT$\to$EN}
&\multicolumn{1}{c}{\bf EN$\to$RU}
&\multicolumn{1}{c}{\bf RU$\to$EN}
&\multicolumn{1}{c}{\bf EN$\to$TR}
&\multicolumn{1}{c}{\bf TR$\to$EN}
&\multicolumn{1}{c}{\bf Avg.}\\ 

\cmidrule(lr){2-16}
\multicolumn{1}{c}{RCSLS} &\multicolumn{1}{c}{57.60} &\multicolumn{1}{c}{56.55} &\multicolumn{1}{c}{42.05} &\multicolumn{1}{c}{41.25} &\multicolumn{1}{c}{66.55} &\multicolumn{1}{c}{63.11} &\multicolumn{1}{c}{37.90} &\multicolumn{1}{c}{35.67} &\multicolumn{1}{c}{64.05} &\multicolumn{1}{c}{61.50} &\multicolumn{1}{c}{49.40} &\multicolumn{1}{c}{48.66} &\multicolumn{1}{c}{39.05} &\multicolumn{1}{c}{37.43} &\multicolumn{1}{c}{50.06} 
\\
\multicolumn{1}{c}{RCSLS + \blice} &\multicolumn{1}{c}{\bf64.00} &\multicolumn{1}{c}{58.95} &\multicolumn{1}{c}{53.60} &\multicolumn{1}{c}{52.60} &\multicolumn{1}{c}{\bf71.75} &\multicolumn{1}{c}{66.17} &\multicolumn{1}{c}{53.15} &\multicolumn{1}{c}{48.92} &\multicolumn{1}{c}{\bf70.50} &\multicolumn{1}{c}{65.79} &\multicolumn{1}{c}{\bf60.45} &\multicolumn{1}{c}{56.26} &\multicolumn{1}{c}{50.35} &\multicolumn{1}{c}{45.74} &\multicolumn{1}{c}{58.44} 
\\
\cmidrule(lr){2-16}
\multicolumn{1}{c}{\vecmap} &\multicolumn{1}{c}{51.00} &\multicolumn{1}{c}{55.24} &\multicolumn{1}{c}{37.75} &\multicolumn{1}{c}{43.51} &\multicolumn{1}{c}{63.10} &\multicolumn{1}{c}{62.75} &\multicolumn{1}{c}{34.05} &\multicolumn{1}{c}{39.08} &\multicolumn{1}{c}{60.40} &\multicolumn{1}{c}{62.17} &\multicolumn{1}{c}{39.65} &\multicolumn{1}{c}{49.35} &\multicolumn{1}{c}{32.05} &\multicolumn{1}{c}{39.24} &\multicolumn{1}{c}{47.81} 
\\
\multicolumn{1}{c}{\vecmap + \blice} &\multicolumn{1}{c}{59.95} &\multicolumn{1}{c}{58.16} &\multicolumn{1}{c}{53.05} &\multicolumn{1}{c}{53.65} &\multicolumn{1}{c}{69.70} &\multicolumn{1}{c}{65.44} &\multicolumn{1}{c}{54.60} &\multicolumn{1}{c}{52.55} &\multicolumn{1}{c}{69.80} &\multicolumn{1}{c}{65.79} &\multicolumn{1}{c}{56.95} &\multicolumn{1}{c}{55.53} &\multicolumn{1}{c}{48.65} &\multicolumn{1}{c}{46.17} &\multicolumn{1}{c}{57.86} 
\\
\cmidrule(lr){2-16}
\multicolumn{1}{c}{C1} &\multicolumn{1}{c}{54.90} &\multicolumn{1}{c}{57.64} &\multicolumn{1}{c}{44.50} &\multicolumn{1}{c}{46.24} &\multicolumn{1}{c}{65.05} &\multicolumn{1}{c}{63.84} &\multicolumn{1}{c}{40.60} &\multicolumn{1}{c}{42.29} &\multicolumn{1}{c}{63.45} &\multicolumn{1}{c}{63.57} &\multicolumn{1}{c}{49.15} &\multicolumn{1}{c}{51.86} &\multicolumn{1}{c}{41.35} &\multicolumn{1}{c}{42.60} &\multicolumn{1}{c}{51.93} 
\\
\multicolumn{1}{c}{C1 + \blice} &\multicolumn{1}{c}{62.75} &\multicolumn{1}{c}{59.68} &\multicolumn{1}{c}{54.25} &\multicolumn{1}{c}{54.02} &\multicolumn{1}{c}{70.75} &\multicolumn{1}{c}{66.48} &\multicolumn{1}{c}{55.40} &\multicolumn{1}{c}{53.55} &\multicolumn{1}{c}{70.05} &\multicolumn{1}{c}{66.10} &\multicolumn{1}{c}{59.25} &\multicolumn{1}{c}{57.41} &\multicolumn{1}{c}{51.05} &\multicolumn{1}{c}{48.14} &\multicolumn{1}{c}{59.21} 
\\
\cmidrule(lr){2-16}
\multicolumn{1}{c}{C2} &\multicolumn{1}{c}{58.05} &\multicolumn{1}{c}{59.31} &\multicolumn{1}{c}{47.15} &\multicolumn{1}{c}{49.97} &\multicolumn{1}{c}{67.55} &\multicolumn{1}{c}{65.39} &\multicolumn{1}{c}{47.85} &\multicolumn{1}{c}{49.13} &\multicolumn{1}{c}{65.25} &\multicolumn{1}{c}{64.65} &\multicolumn{1}{c}{50.80} &\multicolumn{1}{c}{55.21} &\multicolumn{1}{c}{45.05} &\multicolumn{1}{c}{44.46} &\multicolumn{1}{c}{54.99} 
\\
\multicolumn{1}{c}{C2 + \blice} &\multicolumn{1}{c}{63.45} &\multicolumn{1}{c}{\bf60.67} &\multicolumn{1}{c}{\bf55.95} &\multicolumn{1}{c}{\bf55.33} &\multicolumn{1}{c}{70.90} &\multicolumn{1}{c}{\bf67.36} &\multicolumn{1}{c}{\bf57.55} &\multicolumn{1}{c}{\bf55.65} &\multicolumn{1}{c}{70.25} &\multicolumn{1}{c}{\bf66.87} &\multicolumn{1}{c}{60.40} &\multicolumn{1}{c}{\bf58.25} &\multicolumn{1}{c}{\bf52.85} &\multicolumn{1}{c}{\bf48.88} &\multicolumn{1}{c}{\bf60.31} 
\\
\toprule
\rowcolor{Gray}
\multicolumn{1}{c}{[1k] \bf Pairs}  &\multicolumn{1}{c}{\bf EN$\to$DE}  &\multicolumn{1}{c}{\bf DE$\to$EN} &\multicolumn{1}{c}{\bf EN$\to$FI} &\multicolumn{1}{c}{\bf FI$\to$EN}
&\multicolumn{1}{c}{\bf EN$\to$FR}
&\multicolumn{1}{c}{\bf FR$\to$EN}
&\multicolumn{1}{c}{\bf EN$\to$HR}
&\multicolumn{1}{c}{\bf HR$\to$EN}
&\multicolumn{1}{c}{\bf EN$\to$IT}
&\multicolumn{1}{c}{\bf IT$\to$EN}
&\multicolumn{1}{c}{\bf EN$\to$RU}
&\multicolumn{1}{c}{\bf RU$\to$EN}
&\multicolumn{1}{c}{\bf EN$\to$TR}
&\multicolumn{1}{c}{\bf TR$\to$EN}
&\multicolumn{1}{c}{\bf Avg.}\\

\cmidrule(lr){2-16}
\multicolumn{1}{c}{RCSLS} &\multicolumn{1}{c}{46.10} &\multicolumn{1}{c}{48.25} &\multicolumn{1}{c}{28.35} &\multicolumn{1}{c}{28.38} &\multicolumn{1}{c}{56.50} &\multicolumn{1}{c}{55.56} &\multicolumn{1}{c}{22.50} &\multicolumn{1}{c}{22.88} &\multicolumn{1}{c}{55.20} &\multicolumn{1}{c}{53.64} &\multicolumn{1}{c}{35.50} &\multicolumn{1}{c}{36.62} &\multicolumn{1}{c}{23.00} &\multicolumn{1}{c}{24.65} &\multicolumn{1}{c}{38.37} 
\\
\multicolumn{1}{c}{RCSLS + \blice} &\multicolumn{1}{c}{\bf56.50} &\multicolumn{1}{c}{55.97} &\multicolumn{1}{c}{45.90} &\multicolumn{1}{c}{44.56} &\multicolumn{1}{c}{63.65} &\multicolumn{1}{c}{61.87} &\multicolumn{1}{c}{41.10} &\multicolumn{1}{c}{40.03} &\multicolumn{1}{c}{64.45} &\multicolumn{1}{c}{60.83} &\multicolumn{1}{c}{52.25} &\multicolumn{1}{c}{49.40} &\multicolumn{1}{c}{40.20} &\multicolumn{1}{c}{38.55} &\multicolumn{1}{c}{51.09} 
\\
\cmidrule(lr){2-16}
\multicolumn{1}{c}{\vecmap} &\multicolumn{1}{c}{48.25} &\multicolumn{1}{c}{54.25} &\multicolumn{1}{c}{27.75} &\multicolumn{1}{c}{41.30} &\multicolumn{1}{c}{60.30} &\multicolumn{1}{c}{61.25} &\multicolumn{1}{c}{25.50} &\multicolumn{1}{c}{37.56} &\multicolumn{1}{c}{57.45} &\multicolumn{1}{c}{60.88} &\multicolumn{1}{c}{24.80} &\multicolumn{1}{c}{46.31} &\multicolumn{1}{c}{26.55} &\multicolumn{1}{c}{37.11} &\multicolumn{1}{c}{43.52} 
\\
\multicolumn{1}{c}{\vecmap + \blice} &\multicolumn{1}{c}{50.50} &\multicolumn{1}{c}{57.43} &\multicolumn{1}{c}{33.30} &\multicolumn{1}{c}{51.92} &\multicolumn{1}{c}{63.35} &\multicolumn{1}{c}{65.29} &\multicolumn{1}{c}{37.75} &\multicolumn{1}{c}{51.76} &\multicolumn{1}{c}{61.00} &\multicolumn{1}{c}{64.50} &\multicolumn{1}{c}{28.60} &\multicolumn{1}{c}{52.80} &\multicolumn{1}{c}{34.40} &\multicolumn{1}{c}{46.22} &\multicolumn{1}{c}{49.92} 
\\
\cmidrule(lr){2-16}
\multicolumn{1}{c}{C1} &\multicolumn{1}{c}{50.45} &\multicolumn{1}{c}{56.29} &\multicolumn{1}{c}{42.15} &\multicolumn{1}{c}{45.35} &\multicolumn{1}{c}{61.65} &\multicolumn{1}{c}{63.27} &\multicolumn{1}{c}{35.65} &\multicolumn{1}{c}{40.77} &\multicolumn{1}{c}{59.50} &\multicolumn{1}{c}{62.74} &\multicolumn{1}{c}{42.55} &\multicolumn{1}{c}{50.34} &\multicolumn{1}{c}{38.10} &\multicolumn{1}{c}{42.23} &\multicolumn{1}{c}{49.36} 
\\
\multicolumn{1}{c}{C1 + \blice} &\multicolumn{1}{c}{52.50} &\multicolumn{1}{c}{\bf59.36} &\multicolumn{1}{c}{\bf50.95} &\multicolumn{1}{c}{\bf54.02} &\multicolumn{1}{c}{\bf64.40} &\multicolumn{1}{c}{\bf65.75} &\multicolumn{1}{c}{49.30} &\multicolumn{1}{c}{53.34} &\multicolumn{1}{c}{\bf65.05} &\multicolumn{1}{c}{\bf65.12} &\multicolumn{1}{c}{50.80} &\multicolumn{1}{c}{56.21} &\multicolumn{1}{c}{\bf46.55} &\multicolumn{1}{c}{\bf48.40} &\multicolumn{1}{c}{\bf55.84} 
\\
\cmidrule(lr){2-16}
\multicolumn{1}{c}{C2} &\multicolumn{1}{c}{51.00} &\multicolumn{1}{c}{57.17} &\multicolumn{1}{c}{44.45} &\multicolumn{1}{c}{48.34} &\multicolumn{1}{c}{62.05} &\multicolumn{1}{c}{64.25} &\multicolumn{1}{c}{42.35} &\multicolumn{1}{c}{46.82} &\multicolumn{1}{c}{61.35} &\multicolumn{1}{c}{64.03} &\multicolumn{1}{c}{46.15} &\multicolumn{1}{c}{53.17} &\multicolumn{1}{c}{41.30} &\multicolumn{1}{c}{43.56} &\multicolumn{1}{c}{51.86} 
\\
\multicolumn{1}{c}{C2 + \blice} &\multicolumn{1}{c}{51.05} &\multicolumn{1}{c}{58.95} &\multicolumn{1}{c}{50.15} &\multicolumn{1}{c}{53.91} &\multicolumn{1}{c}{63.00} &\multicolumn{1}{c}{65.24} &\multicolumn{1}{c}{\bf50.90} &\multicolumn{1}{c}{\bf54.81} &\multicolumn{1}{c}{62.85} &\multicolumn{1}{c}{64.65} &\multicolumn{1}{c}{\bf52.70} &\multicolumn{1}{c}{\bf56.68} &\multicolumn{1}{c}{46.35} &\multicolumn{1}{c}{47.82} &\multicolumn{1}{c}{55.65} 
\\

\bottomrule
\end{tabular}
}
%\vspace{-1mm}
\caption{BLI scores (P@1$\times100\%$) on the XLING BLI benchmark in supervised and semi-supervised scenarios. We apply the proposed \blice method to all four CLWE-based baselines (i.e., the `baseline + \blice' rows). The scores in \textbf{bold} denote the highest score per column and supervision setup.}
\label{table:main-xling}
%\vspace{-2mm}
\end{center}
\end{table*}

\begin{table*}[!t]
\def\arraystretch{0.99}
\begin{center}
\resizebox{0.999\textwidth}{!}{%
\begin{tabular}{llllllllllllll}
\toprule 

\rowcolor{Gray}
\multicolumn{1}{c}{[1k] \bf Pairs}  &\multicolumn{1}{c}{\bf BG$\to$$*$}&\multicolumn{1}{c}{\bf $*$$\to$BG} &\multicolumn{1}{c}{\bf CA$\to$$*$}&\multicolumn{1}{c}{\bf $*$$\to$CA} &\multicolumn{1}{c}{\bf HE$\to$$*$}&\multicolumn{1}{c}{\bf $*$$\to$HE} &\multicolumn{1}{c}{\bf ET$\to$$*$}&\multicolumn{1}{c}{\bf $*$$\to$ET} &\multicolumn{1}{c}{\bf HU$\to$$*$}&\multicolumn{1}{c}{\bf $*$$\to$HU} &\multicolumn{1}{c}{\bf KA$\to$$*$}&\multicolumn{1}{c}{\bf $*$$\to$KA}
&\multicolumn{1}{c}{\bf Avg.}
\\ 

\cmidrule(lr){2-14}
\multicolumn{1}{c}{RCSLS} &\multicolumn{1}{c}{14.97} &\multicolumn{1}{c}{15.36} &\multicolumn{1}{c}{12.61} &\multicolumn{1}{c}{13.54} &\multicolumn{1}{c}{9.37} &\multicolumn{1}{c}{7.57} &\multicolumn{1}{c}{10.30} &\multicolumn{1}{c}{10.58} &\multicolumn{1}{c}{14.48} &\multicolumn{1}{c}{14.30} &\multicolumn{1}{c}{6.80} &\multicolumn{1}{c}{7.18} &\multicolumn{1}{c}{11.42} 
\\
\multicolumn{1}{c}{RCSLS + \blice} &\multicolumn{1}{c}{30.92} &\multicolumn{1}{c}{31.15} &\multicolumn{1}{c}{26.31} &\multicolumn{1}{c}{26.84} &\multicolumn{1}{c}{19.73} &\multicolumn{1}{c}{17.43} &\multicolumn{1}{c}{23.96} &\multicolumn{1}{c}{25.64} &\multicolumn{1}{c}{28.96} &\multicolumn{1}{c}{28.51} &\multicolumn{1}{c}{17.69} &\multicolumn{1}{c}{18.01} &\multicolumn{1}{c}{24.60} 
\\
\cmidrule(lr){2-14}
\multicolumn{1}{c}{\vecmap} &\multicolumn{1}{c}{32.25} &\multicolumn{1}{c}{31.35} &\multicolumn{1}{c}{26.08} &\multicolumn{1}{c}{32.62} &\multicolumn{1}{c}{26.06} &\multicolumn{1}{c}{24.41} &\multicolumn{1}{c}{26.43} &\multicolumn{1}{c}{23.94} &\multicolumn{1}{c}{30.65} &\multicolumn{1}{c}{33.56} &\multicolumn{1}{c}{22.76} &\multicolumn{1}{c}{18.35} &\multicolumn{1}{c}{27.37} 
\\
\multicolumn{1}{c}{\vecmap + \blice} &\multicolumn{1}{c}{42.15} &\multicolumn{1}{c}{40.60} &\multicolumn{1}{c}{32.39} &\multicolumn{1}{c}{39.20} &\multicolumn{1}{c}{34.14} &\multicolumn{1}{c}{35.54} &\multicolumn{1}{c}{38.05} &\multicolumn{1}{c}{34.75} &\multicolumn{1}{c}{38.38} &\multicolumn{1}{c}{41.20} &\multicolumn{1}{c}{33.90} &\multicolumn{1}{c}{27.70} &\multicolumn{1}{c}{36.50} 
\\
\cmidrule(lr){2-14}
\multicolumn{1}{c}{C1} &\multicolumn{1}{c}{35.96} &\multicolumn{1}{c}{34.97} &\multicolumn{1}{c}{31.41} &\multicolumn{1}{c}{35.04} &\multicolumn{1}{c}{26.37} &\multicolumn{1}{c}{25.69} &\multicolumn{1}{c}{28.06} &\multicolumn{1}{c}{26.85} &\multicolumn{1}{c}{36.45} &\multicolumn{1}{c}{36.38} &\multicolumn{1}{c}{22.34} &\multicolumn{1}{c}{21.65} &\multicolumn{1}{c}{30.10} 
\\
\multicolumn{1}{c}{C1 + \blice} &\multicolumn{1}{c}{47.21} &\multicolumn{1}{c}{44.75} &\multicolumn{1}{c}{41.31} &\multicolumn{1}{c}{43.21} &\multicolumn{1}{c}{36.31} &\multicolumn{1}{c}{37.67} &\multicolumn{1}{c}{41.21} &\multicolumn{1}{c}{40.45} &\multicolumn{1}{c}{45.28} &\multicolumn{1}{c}{44.21} &\multicolumn{1}{c}{34.73} &\multicolumn{1}{c}{35.77} &\multicolumn{1}{c}{41.01} 
\\
\cmidrule(lr){2-14}
\multicolumn{1}{c}{C2} &\multicolumn{1}{c}{40.14} &\multicolumn{1}{c}{38.98} &\multicolumn{1}{c}{35.67} &\multicolumn{1}{c}{39.41} &\multicolumn{1}{c}{30.05} &\multicolumn{1}{c}{29.51} &\multicolumn{1}{c}{33.31} &\multicolumn{1}{c}{33.21} &\multicolumn{1}{c}{39.78} &\multicolumn{1}{c}{38.89} &\multicolumn{1}{c}{25.22} &\multicolumn{1}{c}{24.17} &\multicolumn{1}{c}{34.03} 
\\
\multicolumn{1}{c}{C2 + \blice} &\multicolumn{1}{c}{\bf48.29} &\multicolumn{1}{c}{\bf45.88} &\multicolumn{1}{c}{\bf42.76} &\multicolumn{1}{c}{\bf44.46} &\multicolumn{1}{c}{\bf38.25} &\multicolumn{1}{c}{\bf39.07} &\multicolumn{1}{c}{\bf43.23} &\multicolumn{1}{c}{\bf42.82} &\multicolumn{1}{c}{\bf45.74} &\multicolumn{1}{c}{\bf44.86} &\multicolumn{1}{c}{\bf36.06} &\multicolumn{1}{c}{\bf37.23} &\multicolumn{1}{c}{\bf42.39} 
\\

\bottomrule
\end{tabular}
}
%\vspace{-1.5mm}
\caption{BLI scores (P@1$\times100\%$) on PanLex-BLI. $L\to$$*$ and $*\to$$L$ denote the average scores where $L$ is the source and target language respectively. Detailed results for each language pair are in Appendix~\ref{appendix:panlex}.}
\label{table:panlex-main}
%\vspace{-2mm}
\end{center}
\end{table*}

\begin{table}[!t]
\begin{center}
\def\arraystretch{0.999}
%\resizebox{0.33\textwidth}{!}{%
{\scriptsize
\begin{tabularx}{0.999\linewidth}{l YYY}
\toprule 

\rowcolor{Gray}
\multicolumn{1}{c}{[5k] \textbf{Pairs}}  &\multicolumn{1}{Y}{\bf EN$\to$$*$}  &\multicolumn{1}{Y}{\bf $*$$\to$EN} &\multicolumn{1}{Y}{\bf Avg.} 

\\ \cmidrule(lr){2-4}

\multicolumn{1}{c}{C1}  &\multicolumn{1}{c}{51.29}  &\multicolumn{1}{c}{52.58}  &\multicolumn{1}{c}{51.93}  \\

\multicolumn{1}{c}{C1 + \blice (off-the-shelf)}  &\multicolumn{1}{c}{50.9}  &\multicolumn{1}{c}{52.40}  &\multicolumn{1}{c}{51.65}  \\

\multicolumn{1}{c}{C1 + \blice (w/o Template)}  &\multicolumn{1}{c}{59.81}  &\multicolumn{1}{c}{57.51}  &\multicolumn{1}{c}{58.66}  \\

\multicolumn{1}{c}{C1 + \blice}  &\multicolumn{1}{c}{\bf 60.50}  &\multicolumn{1}{c}{\bf 57.91}  &\multicolumn{1}{c}{\bf 59.21}  \\

\cmidrule(lr){2-4}

\multicolumn{1}{c}{C2}  &\multicolumn{1}{c}{54.53}  &\multicolumn{1}{c}{55.45}  &\multicolumn{1}{c}{54.99}  \\

\multicolumn{1}{c}{C2 + \blice (off-the-shelf)}  &\multicolumn{1}{c}{54.46}  &\multicolumn{1}{c}{55.51}  &\multicolumn{1}{c}{54.99}  \\

\multicolumn{1}{c}{C2 + \blice (w/o Template)}  &\multicolumn{1}{c}{61.31}  &\multicolumn{1}{c}{58.39}  &\multicolumn{1}{c}{59.85}  \\

\multicolumn{1}{c}{C2 + \blice}  &\multicolumn{1}{c}{\bf 61.62}  &\multicolumn{1}{c}{\bf 59.00}  &\multicolumn{1}{c}{\bf 60.31}  \\

\toprule 

\rowcolor{Gray}
\multicolumn{1}{c}{[1k] \textbf{Pairs}}  &\multicolumn{1}{c}{\bf EN$\to$$*$}  &\multicolumn{1}{c}{\bf $*$$\to$EN} &\multicolumn{1}{c}{\bf Avg.} 

\\ \cmidrule(lr){2-4}

\multicolumn{1}{c}{C1}  &\multicolumn{1}{c}{47.15}  &\multicolumn{1}{c}{51.57}  &\multicolumn{1}{c}{49.36}  \\

\multicolumn{1}{c}{C1 + \blice (off-the-shelf)}  &\multicolumn{1}{c}{47.14}  &\multicolumn{1}{c}{51.59}  &\multicolumn{1}{c}{49.37}  \\

\multicolumn{1}{c}{C1 + \blice (w/o Template)}  &\multicolumn{1}{c}{53.96}  &\multicolumn{1}{c}{56.89}  &\multicolumn{1}{c}{55.43}  \\

\multicolumn{1}{c}{C1 + \blice}  &\multicolumn{1}{c}{\bf 54.22}  &\multicolumn{1}{c}{\bf 57.46}  &\multicolumn{1}{c}{\bf 55.84}  \\

\cmidrule(lr){2-4}

\multicolumn{1}{c}{C2}  &\multicolumn{1}{c}{49.81}  &\multicolumn{1}{c}{53.91}  &\multicolumn{1}{c}{51.86}  \\

\multicolumn{1}{c}{C2 + \blice (off-the-shelf)}  &\multicolumn{1}{c}{49.81}  &\multicolumn{1}{c}{53.91}  &\multicolumn{1}{c}{51.86}  \\

\multicolumn{1}{c}{C2 + \blice (w/o Template)}  &\multicolumn{1}{c}{\bf 53.94}  &\multicolumn{1}{c}{57.00}  &\multicolumn{1}{c}{55.47}  \\

\multicolumn{1}{c}{C2 + \blice}  &\multicolumn{1}{c}{53.86}  &\multicolumn{1}{c}{\bf 57.44}  &\multicolumn{1}{c}{\bf 55.65}  \\

\bottomrule
\end{tabularx}
}%
%}

%\vspace{-1.5mm}
\caption{Ablation study. P@1$\times100\%$ scores.}
\label{table:ablation}
\end{center}
%\vspace{-2.5mm}
\end{table}

\rparagraph{Supervised and Semi-Supervised Setups} 
Our results on XLING demonstrate that \blice yields outstanding performance in both supervision setups. In the 5k-setup, the average gain over four CLWE models is $7.76$ P@1 points, and the value is $7.35$ for the 1k-setup. Combining \blice with the two strongest baseline BLI models, C1 and C2, yields average gains of $6.3$ (5k-setup) and $5.14$ points (1k-setup). The baseline BLI scores in the 5k-setup are already much higher than in the 1k-setup, intuitively offering less room for further performance boosts. However, we observe substantial gains with \blice in the 5k-setup across the board, suggesting that \blice effectively leverages the more abundant `gold' bilingual supervision in the 5k-setup, as well as the `silver' supervision derived from the CLWE space, which is more accurate in the 5k-setup than in the 1k-setup. %can benefit from more accurate supervisory signals induced from CLWEs.     

\rparagraph{Compatibility with Different CLWEs} The results indicate that \blice is compatible with all CLWE baselines. The `C2 + \blice' model achieves the highest average score in the XLING (5k) and PanLex-BLI (1k) setups. The `C1 + \blice' variant is the best-performing one in the XLING (1k) setup. Overall, we observe a general trend: \textbf{1)} $\blice$ derives a larger absolute gain when applied to a weaker input CLWE space, but \textbf{2)} starting from a stronger CLWE backbone still yields a stronger `CLWE + \blice' model in terms of the absolute BLI performance.\footnote{There are still some slight deviations from the general trend: e.g., the baseline \vecmap outperforms RCSLS in the XLING (1k) setup on average, but `RCSLS + \blice' surpasses `\vecmap + \blice' by $1.17$ points.}

%E.g., `C1 + \blice' and `C2 + \blice' outperform `RCSLS + \blice' and `\vecmap + \blice' in most cases, since C1 and C2 themselves are stronger CLWEs for BLI. For XLING 1k setups, the average gains on RCSLS and C2 are $12.72$ and $3.79$ points respectively where RCSLS-induces CLWEs are weaker, but `C2 + \blice' still outperforms `RCSLS + \blice'. 

%But this is not universally true. For instance, \vecmap has better average score than RCSLS in XLING 1k setup but `RCSLS + \blice' surpasses `\vecmap + \blice' by $1.17$ points. 

\rparagraph{Performance over Languages} 
The results further indicate that the usefulness of \blice, while observed for all language pairs, is especially pronounced for typologically more distant and lower-resource language pairs. The average gain over four CLWE backbones in the PanLex-BLI (1k) experiments is $10.4$ P@1 points. Directly modeling interaction between two text items (e.g., two words turned into two text templates in our case) which allows them to learn finer-grained `interaction features' \cite{thakur-etal-2021-augmented,geigle-etal-2022-retrieve}, CEs seem especially important in low-resource setups. 

%\fixme{The findings from the previous two works are not specific to low-resource pairs and languages. It seems to relate more with 1k setups, so we may also move this to the "Supervised and Semi-Supervised Setups" paragraph.  Please feel free ignore my comment here if it does not make sense.}

%This aligns with previous studies' finding that CEs achieve strong performance with limited supervision \citep{thakur-etal-2021-augmented}.

\rparagraph{Ablation Study and Further Analysis}
%\subsection{Ablation Study and Further Analysis}
%\label{results:ablation}
We now study the effectiveness of each key component of \blice, basing our analyses on the best-performing baseline CLWEs: C1 and C2. 

%and delve deeper into other analyses. with focus on the SotA C1 and C2 CLWEs only. 

First, \textit{the effectiveness of CE fine-tuning} is indicated by the results in Table~\ref{table:ablation}. Combining off-the-shelf mPLMs with the baseline CLWEs derives no gains at all. We further investigate \textit{the usefulness of templates} (see \S\ref{s:methodology}), with results also provided in Table~\ref{table:ablation}. In general, the scores indicate that templates are not crucial for BLI performance, and simply providing two words to the CE without any extra template also yields very strong BLI performance across the board. We observe only slight average gains in both supervision setups. We also provide average results per each tested template in Appendix~\ref{app:templates}, further suggesting that strong gains are achieved irrespective of the chosen template.

%%adding templates slightly improves the BLI performance but it seems to be a little more useful in 5k setups. 

Further, we study \textit{the impact of the underlying mPLM} on the final BLI performance, with the results summarised in Table~\ref{table:models}. The scores render our \blice method useful with all mPLMs. As expected, the largest XLM-R$_{large}$ model yields the best performance, and we also note a slight edge of XLM-R$_{base}$ over mBERT. %seems to be the best among the other two.

\begin{figure}[t!]
    \centering
    \includegraphics[width=0.97\linewidth]{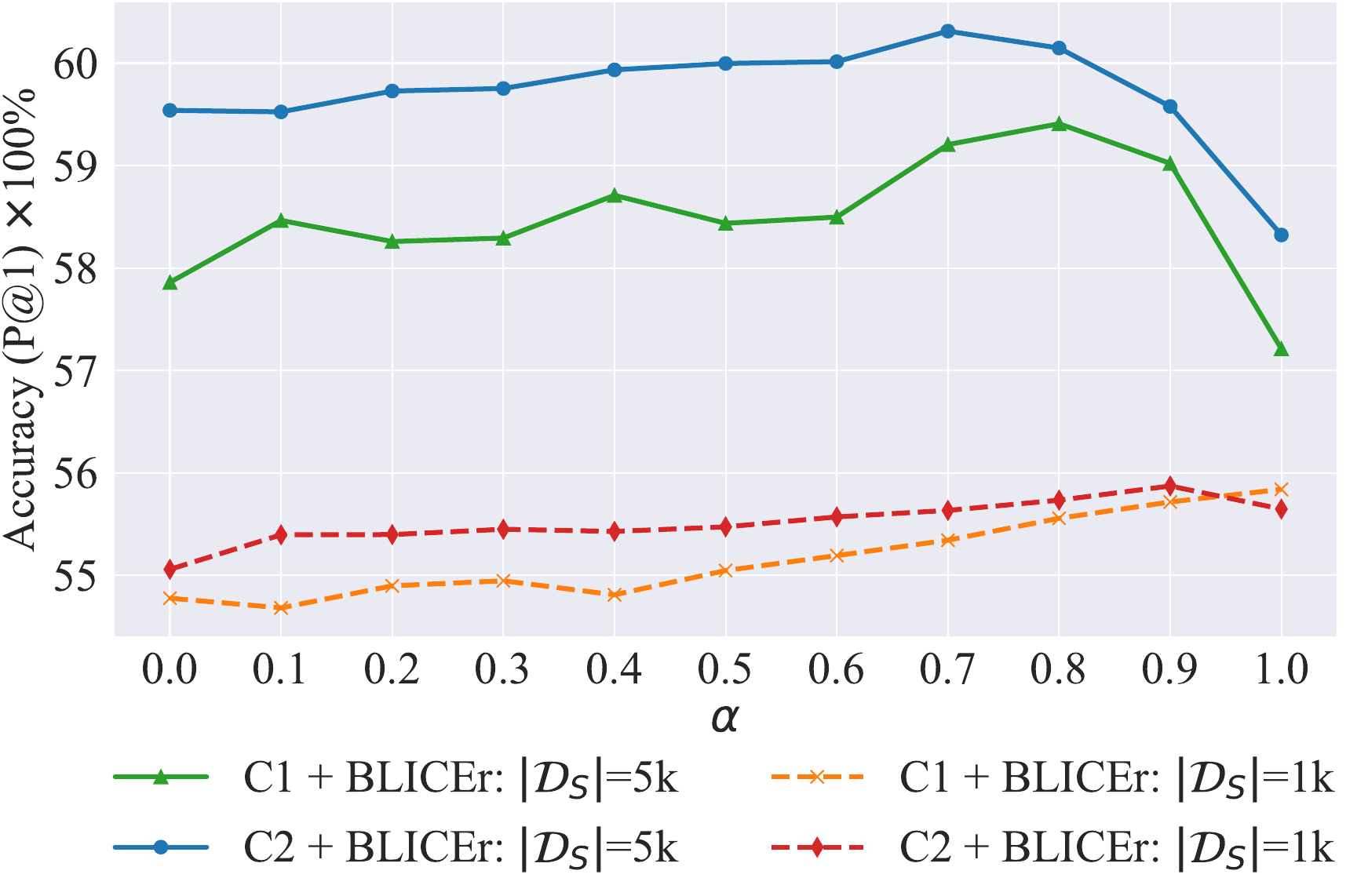}
    %\vspace{-1mm}
    \caption{Average BLI scores on XLING with different values of the polarisation hyper-parameter $\alpha$.}
   % \vspace{-2mm}
    \label{fig:alphas}
%    \vspace{-1.5mm}
\end{figure}

Figure~\ref{fig:alphas} further plots \textit{the impact of polarisation}. First, we note that there are substantial gains with all the $\alpha$ values (cf. Table \ref{table:main-xling}). In the 5k-setup, polarisation achieves a further average gain of 2 points ($\alpha$=$0.7$ versus $\alpha$=$1.0$). In the 1k-setup, it seems that using the original CLWE similarity scores without polarisation (i.e., $\alpha$=$1.0$) yields slightly better results. We attribute this behaviour to potentially noisy `silver' positive pairs in the 1k-setup, which might dilute the gold knowledge from $\mathcal{D}_S$ as the polarisation step might amplify the noise further. Five times more abundant `gold' supervision and more reliable CLWEs in the 5k-setup yield a stronger learning signal for \blice, and this undesirable phenomenon then gets mitigated.
%% \fixme{In 5k setup, we use gold positive pairs and silver neg pairs. Did not use silver positives. `Silver's in other parts of our paper are now all clear.} 

\begin{figure}[t!]
    \centering
    \includegraphics[width=0.97\linewidth]{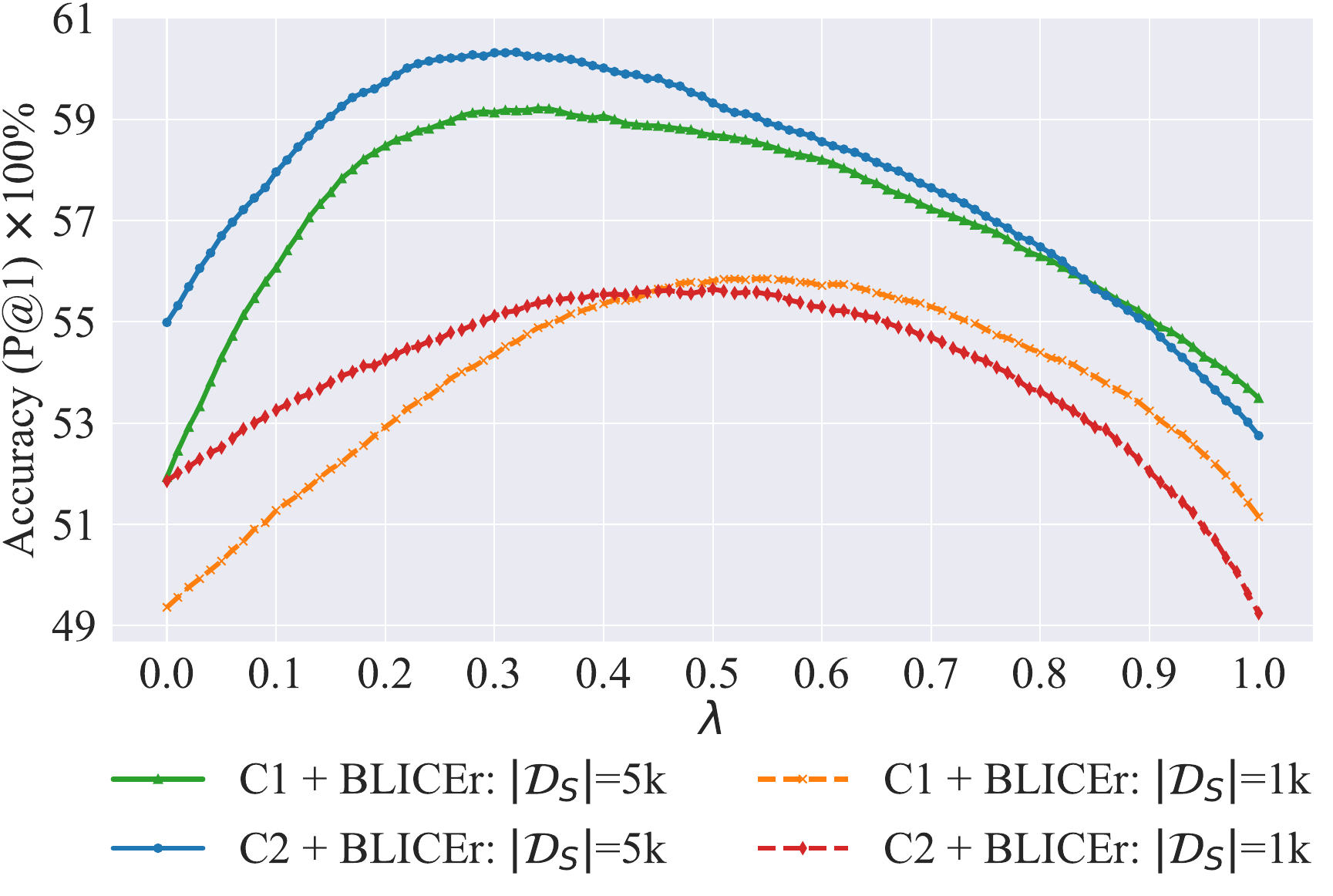}
 %   \vspace{-2mm}
    \caption{Average BLI scores on XLING with different values of the interpolation hyper-parameter $\lambda$. }
    \label{fig:lambdas}
 %   \vspace{-2mm}
\end{figure}

Finally, Figure~\ref{fig:lambdas} demonstrates \textit{the impact of interpolation} of the CE-based scores and the original CLWE scores. The bell-shaped curves across different BLI setups \textbf{1)} clearly indicate the synergistic effect and the usefulness of interpolation across the board, and also \textbf{2)} show that the (near-)optimal $\lambda$ values depend on the amount of supervision. In the 5k-setup, the peak $\lambda$ values put more weight to the original CLWE space ($\lambda$=$0$): this is expected as the starting CLWE space is more accurate when induced with more `gold' supervision, and the CE-based knowledge helps to a lesser extent. The peak $\lambda$ values are higher for the resource-leaner 1k-setup. Figure~\ref{fig:lambdas} also reveals that using only the CE output ($\lambda$=$1$) yields sub-optimal BLI performance, and the true benefit of CE fine-tuning is displayed only in the synergy with the original CLWE space.

%and the model This might show again \blice benefits more from ground-truth supervisory signals. The pseudo positive pairs and the attached `silver' scores in 1k setups are noisy, a polarisation magnify the noise and thus result in sub-optimal performance.   

\begin{table}[!t]
\begin{center}
\def\arraystretch{0.999}
{\scriptsize
%\resizebox{0.33\textwidth}{!}{%
\begin{tabularx}{0.999\linewidth}{l YYY}
\toprule 

\rowcolor{Gray}
\multicolumn{1}{c}{[5k] \textbf{Pairs}}  &\multicolumn{1}{Y}{\bf EN$\to$$*$}  &\multicolumn{1}{Y}{\bf $*$$\to$EN} &\multicolumn{1}{Y}{\bf Avg.} 

\\ \cmidrule(lr){2-4}

\multicolumn{1}{c}{C1}  &\multicolumn{1}{c}{51.29} &\multicolumn{1}{c}{52.58} &\multicolumn{1}{c}{51.93}   \\

\multicolumn{1}{c}{C1 + \blice (mBERT)}  &\multicolumn{1}{c}{55.25} &\multicolumn{1}{c}{54.84} &\multicolumn{1}{c}{55.05}  \\

\multicolumn{1}{c}{C1 + \blice (XLM-R$_{base}$)} &\multicolumn{1}{c}{57.09} &\multicolumn{1}{c}{55.76} &\multicolumn{1}{c}{56.43}   \\

\multicolumn{1}{c}{C1 + \blice (XLM-R$_{large}$)}  &\multicolumn{1}{c}{\bf 60.50} &\multicolumn{1}{c}{\bf 57.91} &\multicolumn{1}{c}{\bf 59.21}  \\

\cmidrule(lr){2-4}

\multicolumn{1}{c}{C2}  &\multicolumn{1}{c}{54.53} &\multicolumn{1}{c}{55.45} &\multicolumn{1}{c}{54.99}   \\

\multicolumn{1}{c}{C2 + \blice (mBERT)}  &\multicolumn{1}{c}{56.29} &\multicolumn{1}{c}{55.82} &\multicolumn{1}{c}{56.06}   \\

\multicolumn{1}{c}{C2 + \blice (XLM-R$_{base}$)}  &\multicolumn{1}{c}{57.59} &\multicolumn{1}{c}{56.57} &\multicolumn{1}{c}{57.08}   \\

\multicolumn{1}{c}{C2 + \blice (XLM-R$_{large}$)}  &\multicolumn{1}{c}{\bf 61.62} &\multicolumn{1}{c}{\bf 59.00} &\multicolumn{1}{c}{\bf 60.31}  \\

\toprule 

\rowcolor{Gray}
\multicolumn{1}{c}{[1k] \textbf{Pairs}}  &\multicolumn{1}{c}{\bf EN$\to$$*$}  &\multicolumn{1}{c}{\bf $*$$\to$EN} &\multicolumn{1}{c}{\bf Avg.} 

\\ \cmidrule(lr){2-4}

\multicolumn{1}{c}{C1}  &\multicolumn{1}{c}{47.15} &\multicolumn{1}{c}{51.57} &\multicolumn{1}{c}{49.36} \\

\multicolumn{1}{c}{C1 + \blice (mBERT)}  &\multicolumn{1}{c}{50.36} &\multicolumn{1}{c}{54.25} &\multicolumn{1}{c}{52.30}  \\

\multicolumn{1}{c}{C1 + \blice (XLM-R$_{base}$)}  &\multicolumn{1}{c}{51.46} &\multicolumn{1}{c}{55.17} &\multicolumn{1}{c}{53.31}  \\

\multicolumn{1}{c}{C1 + \blice (XLM-R$_{large}$)}  &\multicolumn{1}{c}{\bf 54.22} &\multicolumn{1}{c}{\bf 57.46} &\multicolumn{1}{c}{\bf 55.84} \\

\cmidrule(lr){2-4}

\multicolumn{1}{c}{C2}  &\multicolumn{1}{c}{49.81} &\multicolumn{1}{c}{53.91} &\multicolumn{1}{c}{51.86} \\

\multicolumn{1}{c}{C2 + \blice (mBERT)}  &\multicolumn{1}{c}{50.28} &\multicolumn{1}{c}{54.58} &\multicolumn{1}{c}{52.43} \\

\multicolumn{1}{c}{C2 + \blice (XLM-R$_{base}$)}  &\multicolumn{1}{c}{51.19} &\multicolumn{1}{c}{55.38} &\multicolumn{1}{c}{53.28}  \\

\multicolumn{1}{c}{C2 + \blice (XLM-R$_{large}$)} &\multicolumn{1}{c}{\bf 53.86} &\multicolumn{1}{c}{\bf 57.44} &\multicolumn{1}{c}{\bf 55.65} \\

\bottomrule
\end{tabularx}
%}
}%

%\vspace{-1.5mm}
\caption{\blice based on different pretrained LMs.}
\label{table:models}
\end{center}
%\vspace{-1.5mm}
\end{table}

\rparagraph{Unsupervised and Zero-Shot Setups}
While this paper mainly focuses on arguably more practical supervised and semi-supervised settings,\footnote{For instance, \citet{vulic-etal-2019-really} empirically prove that using even a small amount of supervision (e.g., 200, 500 or 1,000 word translation pairs) always outperforms fully unsupervised BLI methods, while \citet{Artetxe:2020acl} and \citet{Wang:2022acl} discuss that at least some word translation pairs such as PanLex dictionaries \cite{Kamholz2014panlex} are available for thousands of the world's languages.} we also conduct preliminary investigations of \blice in \textit{fully unsupervised} and \textit{zero-shot} settings, where \textit{no direct bilingual supervision} between the source and the target is assumed: see Appendix~\ref{s:unsup} for the overview of the experimental setup. The results in these extreme settings, provided in Appendix~\ref{s:unsup}, further validate the usefulness of \blice as a post-processing method: it again yields substantial and consistent gains when applied to the backbone fully unsupervised \vecmap CLWE space.

\section{Conclusion}
\label{s:conclusion}
We presented \blice, a simple and effective \textit{post-hoc} reranking method for improved bilingual lexicon induction (BLI). \blice is applicable to any underlying cross-lingual word embedding (CLWE) space. It is based on fine-tuning multilingual pretrained language models into BLI-oriented cross-encoders with a limited amount of direct bilingual supervision (i.e., seed word translation pairs). At BLI inference, the \blice output refines cross-lingual word similarities from the underlying CLWE space. We conducted extensive empirical studies covering a total of $352$ supervised and semi-supervised BLI setups, and observe substantial gains against representative and strong BLI baselines across the board. We also performed a series of ablation studies and validated the unsupervised and zero-shot capabilities of \blice. In future research we plan to experiment with other multilingual language models \cite{He:2021deberta} and their ensembles, and we will extend the work to other languages and multilingual lexical tasks. 

%% the supervision signal for fine-tuning spans gold bilingual supervision (i.e., seed word translation pairs), augmented with `silver' supervision from the original CLWE space, and hard negative pairs from the CLWE space. 

\section*{Acknowledgements}
\label{s:acknowledgements}
{\scriptsize\euflag} This work has been partially supported by the ERC PoC Grant MultiConvAI (no. 957356) and a research donation from Huawei. YL and FL are supported by Grace $\&$ Thomas C. H. Chan Cambridge International Scholarship. IV is also supported by a personal Royal Society University Research Fellowship.

%\section*{Ethics Statement}
%\label{s:ethics}
%\input{8_ethics}

\section*{Limitations}
\label{s:limitations}
%\begin{itemize}
There are almost $7,000$ languages worldwide \cite{ethnologue}. However, publicly available fastText word embeddings currently only cover 294 languages,\footnote{\url{https://fasttext.cc/docs/en/pretrained-vectors.html}} mBERT supports only $104$ languages,\footnote{\url{https://github.com/google-research/bert}} and XLM-R only $100$.\footnote{\url{https://github.com/facebookresearch/XLM}} More effort is needed towards building bilingual dictionaries and language technology tools for under-represented and low-resource languages. Researchers, in the future, may also consider to develop techniques to address low-resource languages even without enough monolingual data for pretraining language models. Our work does not extend the scope to additional languages, and is by proxy also constrained by the current limitations of the underlying models such as fastText, mBERT, and XLM-R.

Some of the existing established BLI datasets were built with publicly available translation tools such as Google Translate plus some simple \textit{post-hoc} refinements \cite{conneau2017word,glavas-etal-2019-properly}. There are occasionally noisy data points in the supposedly `gold standard' datasets \cite{kementchedjhieva-etal-2019-lost}, they are typically not fully adapted to languages with more productive morphosyntactic systems \cite{czarnowska-etal-2019-dont}, and the control of synonyms and polysemy is difficult. While these evaluation data deficiencies do not impact relative comparisons between BLI models, for real-world applications gold standard ground-truth data of higher quality are needed for a vast number of language pairs. Their careful creation and annotation should involve native speakers of (low-resource) target languages, bilingual speakers and linguists. %should be involved in data annotation. 
%\end{itemize}

\bibliography{anthology,custom}
\bibliographystyle{acl_natbib}

\clearpage
\appendix

\section{Languages in Experiments}

\begin{table}[ht!]
\begin{center}
\resizebox{0.9\linewidth}{!}{%
\begin{tabular}{llll}
\toprule 
\rowcolor{Gray}
\multicolumn{1}{l}{\bf Family}  &\multicolumn{1}{l}{\bf Language} &\multicolumn{1}{l}{\bf $[L_w]$}   &\multicolumn{1}{l}{\bf Code}\\
\hline
\multirow{1}{*}{Afro-Asiatic}&\multicolumn{1}{l}{Hebrew}&\multicolumn{1}{l}{\inlinegraphics{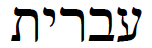}} &\multicolumn{1}{l}{\textsc{he}}\\
\hline
\multirow{2}{*}{Germanic}&\multicolumn{1}{l}{English} &\multicolumn{1}{l}{english} &\multicolumn{1}{l}{\textsc{en}}\\
&\multicolumn{1}{l}{German}  &\multicolumn{1}{l}{deutsch}&\multicolumn{1}{l}{\textsc{de}}\\
\hline
\multirow{1}{*}{Kartvelian}&\multicolumn{1}{l}{Georgian} &\multicolumn{1}{l}{\inlinegraphics{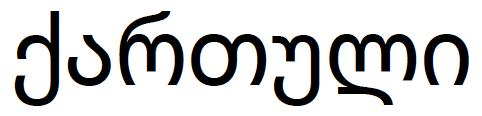}} &\multicolumn{1}{l}{\textsc{ka}}\\
\hline
\multirow{3}{*}{Romance}&\multicolumn{1}{l}{Catalan}&\multicolumn{1}{l}{català}&\multicolumn{1}{l}{\textsc{ca}}\\&\multicolumn{1}{l}{French}&\multicolumn{1}{l}{français}&\multicolumn{1}{l}{\textsc{fr}}\\
&\multicolumn{1}{l}{Italian}  &\multicolumn{1}{l}{italiano}&\multicolumn{1}{l}{\textsc{it}}\\
\hline
\multirow{3}{*}{Slavic}&\multicolumn{1}{l}{Bulgarian}&\multicolumn{1}{l}{\inlinegraphics{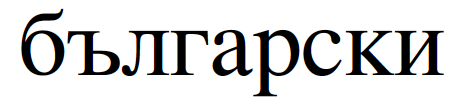}}&\multicolumn{1}{l}{\textsc{bg}}\\&\multicolumn{1}{l}{Croatian}&\multicolumn{1}{l}{hrvatski}&\multicolumn{1}{l}{\textsc{hr}}\\&\multicolumn{1}{l}{Russian}  &\multicolumn{1}{l}{\inlinegraphics{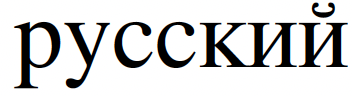}}&\multicolumn{1}{l}{\textsc{ru}}\\
\hline
\multirow{1}{*}{Turkic}&\multicolumn{1}{l}{Turkish} &\multicolumn{1}{l}{türkçe}&\multicolumn{1}{l}{\textsc{tr}}\\
\hline
\multirow{3}{*}{Uralic}&\multicolumn{1}{l}{Estonian}&\multicolumn{1}{l}{eesti}&\multicolumn{1}{l}{\textsc{et}}\\&\multicolumn{1}{l}{Finnish}&\multicolumn{1}{l}{suomi}&\multicolumn{1}{l}{\textsc{fi}}\\&\multicolumn{1}{l}{Hungarian }&\multicolumn{1}{l}{magyar}&\multicolumn{1}{l}{\textsc{hu}}\\
\bottomrule
\end{tabular}
}

\caption{Languages involved in our experiments, categorized by language family. We also show their ISO 639-1 codes and $[L_w]$ used in some text templates (see \S\ref{s:methodology} and Appendix~\ref{app:templates}).}
\label{table:languages_new}
\end{center}
\end{table}

\section{Text Templates}
\label{app:templates}
We experiment with $16$ different templates as follows. Among the $16$ templates, $T_{1}-T_{4}$ are $4$ basic templates. $T_{5}-T_{8}$ add a quotation mark (i.e., `') to each $[w]$. $T_{9}-T_{12}$ add a full stop (i.e., `.') at the end of the template, and $T_{13}-T_{16}$ append the exclamation mark (i.e., `!') to each sequence.
\begin{center}
\footnotesize{
\begin{tcolorbox}[colback=gray!10,%gray background
                  colframe=black,% black frame colour
                  width=5.6cm,% Use 8cm total width,
                  arc=1mm, auto outer arc,
                  boxrule=0.6pt,
                 ]
$T_{1}$:\;  $[w]$ \\
$T_{2}$:\;  $\text{the word}\ [w]$ \\
$T_{3}$:\;  $[w]\ ([L_w])$ \\
$T_{4}$:\;  $\text{the word}\ [w]\ \text{in}\ [L_w]$ \\
$T_{5}$:\;  $\text{`}[w]\text{'}$ \\
$T_{6}$:\;  $\text{the word}\ \text{`}[w]\text{'}$ \\
$T_{7}$:\;  $\text{`}[w]\text{'}\ ([L_w])$ \\
$T_{8}$:\;  $\text{the word `}[w]\text{' in}\ [L_w]$ \\
$T_{9}$:\;  $[w]\text{.}$ \\
$T_{10}$: $\text{the word}\ [w].$ \\
$T_{11}$: $[w]\ ([L_w]).$ \\
$T_{12}$: $\text{the word}\ [w]\ \text{in}\ [L_w].$ \\
$T_{13}$: $[w]!$ \\
$T_{14}$: $\text{the word}\ [w]!$ \\
$T_{15}$: $[w]\ ([L_w])!$ \\
$T_{16}$: $\text{the word}\ [w]\ \text{in}\ [L_w]!$
\end{tcolorbox}
}
\end{center}

An analysis supplementing the main analysis in the main paper (see \S\ref{s:results}) shows the average results with each template; the model variant is `C2 + \blice' evaluated on XLING (5k) covering 14 BLI directions. The results are summarized in Table~\ref{table:templates}.

First, among $T_{1}-T_{4}$, there is a very minor variation in the results, where $T_{3}$ seems to be slightly better than the other three templates. Second, adding quotation marks ($T_{5}-T_{8}$) results in decreased BLI performance when compared to the basic templates (or using no template at all). Third, adding a full stop or an exclamation mark ($T_{9}-T_{16}$) does not have any real impact on the results. We again emphasize \textbf{1)} that in our main experiments we pick the template $T_{15}$ that achieves best performance on a single language pair (\textsc{en},\textsc{tr}), which we also use to tune all the hyper-parameters, but \textbf{2)} not using any template (effectively using the template $T_1$) also yields very strong results across the board (see also the main paper). 

%We do not know the results in Table \ref{table:templates} when conducting main experiments.

\begin{table}[t!]
\begin{center}
\resizebox{0.3\textwidth}{!}{%
\begin{tabular}{llll}

\toprule 

\rowcolor{Gray}
\multicolumn{1}{c}{\bf $T_{1}$}  &\multicolumn{1}{c}{\bf $T_{2}$}  &\multicolumn{1}{c}{\bf $T_{3}$} &\multicolumn{1}{c}{\bf $T_{4}$}\\

\multicolumn{1}{c}{59.85}  &\multicolumn{1}{c}{59.59}  &\multicolumn{1}{c}{60.15}  &\multicolumn{1}{c}{60.04}  \\

\toprule 

\rowcolor{Gray}
\multicolumn{1}{c}{\bf $T_{5}$}  &\multicolumn{1}{c}{\bf $T_{6}$}  &\multicolumn{1}{c}{\bf $T_{7}$} &\multicolumn{1}{c}{\bf $T_{8}$}\\

\multicolumn{1}{c}{58.60}  &\multicolumn{1}{c}{58.38}  &\multicolumn{1}{c}{58.34}  &\multicolumn{1}{c}{58.47}  \\

\toprule 

\rowcolor{Gray}
\multicolumn{1}{c}{\bf $T_{9}$}  &\multicolumn{1}{c}{\bf $T_{10}$}  &\multicolumn{1}{c}{\bf $T_{11}$} &\multicolumn{1}{c}{\bf $T_{12}$}\\

\multicolumn{1}{c}{59.16}  &\multicolumn{1}{c}{59.80}  &\multicolumn{1}{c}{60.12}  &\multicolumn{1}{c}{59.88}  \\

\toprule 

\rowcolor{Gray}
\multicolumn{1}{c}{\bf $T_{13}$}  &\multicolumn{1}{c}{\bf $T_{14}$}  &\multicolumn{1}{c}{\bf $T_{15}$} &\multicolumn{1}{c}{\bf $T_{16}$}\\

\multicolumn{1}{c}{59.85}  &\multicolumn{1}{c}{59.82}  &\multicolumn{1}{c}{\bf 60.31}  &\multicolumn{1}{c}{59.97}  \\

\bottomrule

\end{tabular}
}

\caption{`C2 + \blice' with different templates. P@1$\times100\%$ scores (average over all 14 BLI directions in the XLING benchmark (5k pairs).}
\label{table:templates}
\end{center}
\end{table}

\begin{table*}[!ht]
\begin{center}
\def\arraystretch{0.999}
\resizebox{0.87\textwidth}{!}{%
{\footnotesize
\begin{tabular}{l ccccccc}
\toprule 

\rowcolor{Gray}
\multicolumn{1}{c}{}  &\multicolumn{1}{c}{\bf DE$\to$TR}  &\multicolumn{1}{c}{\bf TR$\to$DE} &\multicolumn{1}{c}{\bf FI$\to$HR}  &\multicolumn{1}{c}{\bf HR$\to$FI} &\multicolumn{1}{c}{\bf IT$\to$FR}  &\multicolumn{1}{c}{\bf FR$\to$IT} &\multicolumn{1}{c}{\bf Avg.}

\\ \cmidrule(lr){2-8}

\multicolumn{1}{c}{\vecmap}  &\multicolumn{1}{c}{23.79} &\multicolumn{1}{c}{26.46} &\multicolumn{1}{c}{28.80} &\multicolumn{1}{c}{27.72} &\multicolumn{1}{c}{\bf65.22} &\multicolumn{1}{c}{63.42} &\multicolumn{1}{c}{39.24}  \\

\multicolumn{1}{c}{+\blice (unsuper)} &\multicolumn{1}{c}{29.53} &\multicolumn{1}{c}{\bf35.78} &\multicolumn{1}{c}{\bf37.78} &\multicolumn{1}{c}{\bf37.19} &\multicolumn{1}{c}{64.44} &\multicolumn{1}{c}{64.46} &\multicolumn{1}{c}{44.86} \\

%\toprule 

%\rowcolor{Gray}
%\multicolumn{1}{c}{\bf Zero-shot}   &\multicolumn{1}{c}{\bf DE$\to$TR}  &\multicolumn{1}{c}{\bf TR$\to$DE} &\multicolumn{1}{c}{\bf FI$\to$HR}  &\multicolumn{1}{c}{\bf HR$\to$FI} &\multicolumn{1}{c}{\bf IT$\to$FR}  &\multicolumn{1}{c}{\bf FR$\to$IT}
%&\multicolumn{1}{c}{\bf Avg.}

%\\ \cmidrule(lr){2-8}

%\multicolumn{1}{c}{\vecmap}  &\multicolumn{1}{c}{23.79} &\multicolumn{1}{c}{26.46} &\multicolumn{1}{c}{28.80} &\multicolumn{1}{c}{27.72} &\multicolumn{1}{c}{\bf65.22} &\multicolumn{1}{c}{63.42} &\multicolumn{1}{c}{39.24}  \\

\multicolumn{1}{c}{+\blice (zero-shot)}  &\multicolumn{1}{c}{\bf33.54}  &\multicolumn{1}{c}{35.68}  &\multicolumn{1}{c}{35.16} &\multicolumn{1}{c}{37.09} &\multicolumn{1}{c}{63.10} &\multicolumn{1}{c}{\bf64.82} &\multicolumn{1}{c}{\bf44.90} \\

\bottomrule
\end{tabular}
}%
}

%\vspace{-1.5mm}
\caption{\blice in unsupervised and zero-shot setups.}
\label{table:unsupzeroshot}
\end{center}
%\vspace{-2.5mm}
\end{table*}

\section{Unsupervised and Zero-shot Setups}
\label{s:unsup}
This paper mainly focuses on the more practical supervised and semi-supervised settings, but as mentioned in the main paper (see the last paragraph of \S\ref{s:results}) we also conduct preliminary investigations of \blice in \textit{fully unsupervised} and \textit{zero-shot} settings, where no direct bilingual supervision between the source and the target is assumed. We rely on the unsupervised variant of \vecmap, a strong unsupervised BLI method \cite{glavas-etal-2019-properly}, as the CLWE backbone for \blice: among our four CLWE baselines, \vecmap is the only one that supports \textit{fully unsupervised} BLI.

We consider BLI tasks between several (randomly sampled) language pairs from the XLING benchmark that do not include English as one of the languages, and dispose of any direct bilingual supervision. \textbf{First}, the \textit{fully unsupervised} setup is in fact a variant of our semi-supervised setup: for CE fine-tuning we now use only `silver' word translation pairs obtained from the unsupervised CLWE space. We rely on the same hyper-parameters as in the semi-supervised setups (see \S\ref{s:experimental}). \textbf{Second}, in the \textit{zero-shot} setup, while translating from the language $L_x$ to $L_y$, we assume that we only possess sets of word translations for the language pairs (\textsc{en}, $L_x$) and (\textsc{en}, $L_y$). This experiment aims to verify if \blice can effectively leverage the inherent multilinguality of the underlying mPLMs.\footnote{The input domains during training and evaluation are totally different. The zero-shot capability of \blice may show that the zero-shot training with (\textsc{en}, $L_x$) and (\textsc{en}, $L_y$) input can expose the word translation knowledge from mPLMs for BLI between the pair ($L_x$, $L_y$).} In particular, we assume 2k seed word pairs for (\textsc{en}, $L_x$) and another 2k pairs for (\textsc{en}, $L_y$).\footnote{We randomly sample the 2k pairs from the respective 5k training sets in XLING. We additionally ensure that there is no overlap of English words between the two sets. This constraint prevents naively deriving word pairs between $L_x$ and $L_y$ from the two seed dictionaries via using the same word as the pivot.} For CE fine-tuning, we then use the $4k$ positive word pairs, together with negative pairs retrieved from the two CLWE spaces,\footnote{The negatives are derived from the semi-supervised C2-based CLWE spaces which are based on the provided dictionaries of 2k pairs. We rely on semi-supervised C2 only for deriving bilingual supervision for CE fine-tune, where (\textsc{en}, $L_x$) and (\textsc{en}, $L_y$) input is fed to the CE. At BLI inference, we use instead cross-lingual word similarity scores obtained from unsupervised \vecmap for ($L_x$, $L_y$).} without any augmentation with `silver' translation pairs. We adopt the hyper-parameter values from the supervised setup (see \S\ref{s:experimental} again).

The results in Table~\ref{table:unsupzeroshot} demonstrate that \blice is also effective in unsupervised and zero-shot settings, yielding substantial gains over the unsupervised \vecmap baseline on average. The only exception is the highly similar language pair \textsc{it}-\textsc{fr}, where the baseline CLWE model already strikes extremely high P@1 performance. Importantly, the results in zero-shot setups validate that \blice implicitly benefits from the multilingual information stored in the underlying XLM-R$_{large}$ model.

In addition, we also show `\vecmap + \blice' results on XLING EN$\to$$*$ and $*$$\to$EN in \textit{fully unsupervised} setups. Table~\ref{table:appendix-unsupervised} presents \blice results where the unsupervised \vecmap model is used as the CLWE backbone. Note that the hyper-parameters are also tuned on the language pair (\textsc{en}, \textsc{tr}). Consequently, we point out that the `\vecmap + \blice' results below in Table~\ref{table:appendix-unsupervised} for the \textsc{en}$\to$\textsc{tr} and \textsc{tr}$\to$\textsc{en} are in fact not unsupervised (we include them only for completeness). 

\begin{table*}[ht!]
\begin{center}
\resizebox{0.999\textwidth}{!}{%
\begin{tabular}{llllllllllllllll}
\toprule 

\rowcolor{Gray}
\multicolumn{1}{c}{}  &\multicolumn{1}{c}{\bf EN$\to$DE}  &\multicolumn{1}{c}{\bf DE$\to$EN} &\multicolumn{1}{c}{\bf EN$\to$FI} &\multicolumn{1}{c}{\bf FI$\to$EN}
&\multicolumn{1}{c}{\bf EN$\to$FR}
&\multicolumn{1}{c}{\bf FR$\to$EN}
&\multicolumn{1}{c}{\bf EN$\to$HR}
&\multicolumn{1}{c}{\bf HR$\to$EN}
&\multicolumn{1}{c}{\bf EN$\to$IT}
&\multicolumn{1}{c}{\bf IT$\to$EN}
&\multicolumn{1}{c}{\bf EN$\to$RU}
&\multicolumn{1}{c}{\bf RU$\to$EN}
&\multicolumn{1}{c}{\bf EN$\to$TR}
&\multicolumn{1}{c}{\bf TR$\to$EN}
&\multicolumn{1}{c}{\bf Avg.}\\ 

\cmidrule(lr){2-16}
\multicolumn{1}{c}{\vecmap} &\multicolumn{1}{c}{48.45} &\multicolumn{1}{c}{54.51} &\multicolumn{1}{c}{28.15} &\multicolumn{1}{c}{41.04} &\multicolumn{1}{c}{60.10} &\multicolumn{1}{c}{61.51} &\multicolumn{1}{c}{24.10} &\multicolumn{1}{c}{36.30} &\multicolumn{1}{c}{57.40} &\multicolumn{1}{c}{60.78} &\multicolumn{1}{c}{25.10} &\multicolumn{1}{c}{46.41} &\multicolumn{1}{c}{26.50} &\multicolumn{1}{c}{36.90} &\multicolumn{1}{c}{43.37} 
\\
\multicolumn{1}{c}{\vecmap + \blice (unsuper)} &\multicolumn{1}{c}{\bf51.30} &\multicolumn{1}{c}{\bf57.07} &\multicolumn{1}{c}{\bf36.95} &\multicolumn{1}{c}{\bf52.92} &\multicolumn{1}{c}{\bf63.35} &\multicolumn{1}{c}{\bf64.20} &\multicolumn{1}{c}{\bf36.45} &\multicolumn{1}{c}{\bf50.34} &\multicolumn{1}{c}{\bf62.05} &\multicolumn{1}{c}{\bf64.03} &\multicolumn{1}{c}{\bf29.15} &\multicolumn{1}{c}{\bf52.65} &\multicolumn{1}{c}{\bf37.05} &\multicolumn{1}{c}{\bf46.11} &\multicolumn{1}{c}{\bf50.26} 
\\

\bottomrule
\end{tabular}
}

\caption{BLI scores (P@1$\times100\%$) on the XLING benchmark, EN$\to$$*$ and $*$$\to$EN unsupervised BLI tasks. Unsupervised \vecmap is used as the CLWE backbone, and we use only `silver' word translation pairs for cross-encoder fine-tuning (see Appendix~\ref{s:unsup}).}
\label{table:appendix-unsupervised}
\end{center}
\end{table*}

\section{BLI Results on PanLex-BLI for Individual Language Pairs}
\label{appendix:panlex}

In Table~\ref{table:appendix-panlex}, we present full BLI results per each PanLex-BLI language pair, while the results in the main paper are aggregated over a particular source or target language (see Table~\ref{table:panlex-main}).

\begin{table*}[ht]
\begin{center}
\resizebox{0.99\textwidth}{!}{%
\begin{tabular}{llllllllllllllll}
\toprule 

\rowcolor{Gray}
\multicolumn{1}{c}{[1k] \bf Pairs - First Half}  &\multicolumn{1}{c}{\bf BG$\to$CA}  &\multicolumn{1}{c}{\bf BG$\to$HE} &\multicolumn{1}{c}{\bf BG$\to$ET} &\multicolumn{1}{c}{\bf BG$\to$HU}
&\multicolumn{1}{c}{\bf BG$\to$KA}
&\multicolumn{1}{c}{\bf CA$\to$HE}
&\multicolumn{1}{c}{\bf CA$\to$ET}
&\multicolumn{1}{c}{\bf CA$\to$HU}
&\multicolumn{1}{c}{\bf CA$\to$KA}
&\multicolumn{1}{c}{\bf HE$\to$ET}
&\multicolumn{1}{c}{\bf HE$\to$HU}
&\multicolumn{1}{c}{\bf HE$\to$KA}
&\multicolumn{1}{c}{\bf ET$\to$HU}
&\multicolumn{1}{c}{\bf ET$\to$KA}
&\multicolumn{1}{c}{\bf HU$\to$KA}\\

\cmidrule(lr){2-16}
\multicolumn{1}{c}{RCSLS} &\multicolumn{1}{c}{18.40} &\multicolumn{1}{c}{10.86} &\multicolumn{1}{c}{14.92} &\multicolumn{1}{c}{19.44} &\multicolumn{1}{c}{11.25} &\multicolumn{1}{c}{9.36} &\multicolumn{1}{c}{10.39} &\multicolumn{1}{c}{18.12} &\multicolumn{1}{c}{6.66} &\multicolumn{1}{c}{5.51} &\multicolumn{1}{c}{10.77} &\multicolumn{1}{c}{3.47} &\multicolumn{1}{c}{15.57} &\multicolumn{1}{c}{6.55} &\multicolumn{1}{c}{7.99} 
\\
\multicolumn{1}{c}{RCSLS + \blice} &\multicolumn{1}{c}{34.29} &\multicolumn{1}{c}{23.48} &\multicolumn{1}{c}{32.95} &\multicolumn{1}{c}{36.40} &\multicolumn{1}{c}{27.51} &\multicolumn{1}{c}{20.34} &\multicolumn{1}{c}{26.66} &\multicolumn{1}{c}{31.33} &\multicolumn{1}{c}{19.86} &\multicolumn{1}{c}{16.14} &\multicolumn{1}{c}{24.57} &\multicolumn{1}{c}{3.47} &\multicolumn{1}{c}{32.14} &\multicolumn{1}{c}{18.36} &\multicolumn{1}{c}{20.85} 
\\
\cmidrule(lr){2-16}
\multicolumn{1}{c}{\vecmap} &\multicolumn{1}{c}{39.66} &\multicolumn{1}{c}{30.80} &\multicolumn{1}{c}{27.88} &\multicolumn{1}{c}{38.77} &\multicolumn{1}{c}{24.13} &\multicolumn{1}{c}{24.87} &\multicolumn{1}{c}{22.21} &\multicolumn{1}{c}{35.47} &\multicolumn{1}{c}{14.29} &\multicolumn{1}{c}{17.66} &\multicolumn{1}{c}{33.56} &\multicolumn{1}{c}{13.61} &\multicolumn{1}{c}{35.60} &\multicolumn{1}{c}{17.80} &\multicolumn{1}{c}{21.91} 
\\
\multicolumn{1}{c}{\vecmap + \blice} &\multicolumn{1}{c}{41.88} &\multicolumn{1}{c}{44.96} &\multicolumn{1}{c}{41.13} &\multicolumn{1}{c}{46.64} &\multicolumn{1}{c}{36.13} &\multicolumn{1}{c}{31.67} &\multicolumn{1}{c}{32.88} &\multicolumn{1}{c}{39.67} &\multicolumn{1}{c}{19.58} &\multicolumn{1}{c}{28.68} &\multicolumn{1}{c}{39.75} &\multicolumn{1}{c}{24.41} &\multicolumn{1}{c}{44.53} &\multicolumn{1}{c}{27.87} &\multicolumn{1}{c}{30.52} 
\\
\cmidrule(lr){2-16}
\multicolumn{1}{c}{C1} &\multicolumn{1}{c}{41.88} &\multicolumn{1}{c}{33.92} &\multicolumn{1}{c}{33.47} &\multicolumn{1}{c}{41.49} &\multicolumn{1}{c}{29.02} &\multicolumn{1}{c}{30.39} &\multicolumn{1}{c}{26.72} &\multicolumn{1}{c}{38.78} &\multicolumn{1}{c}{23.02} &\multicolumn{1}{c}{16.54} &\multicolumn{1}{c}{35.34} &\multicolumn{1}{c}{11.45} &\multicolumn{1}{c}{40.23} &\multicolumn{1}{c}{17.43} &\multicolumn{1}{c}{27.33} 
\\
\multicolumn{1}{c}{C1 + \blice} &\multicolumn{1}{c}{45.39} &\multicolumn{1}{c}{46.08} &\multicolumn{1}{c}{49.19} &\multicolumn{1}{c}{49.54} &\multicolumn{1}{c}{45.86} &\multicolumn{1}{c}{40.62} &\multicolumn{1}{c}{41.29} &\multicolumn{1}{c}{44.31} &\multicolumn{1}{c}{36.17} &\multicolumn{1}{c}{30.82} &\multicolumn{1}{c}{41.41} &\multicolumn{1}{c}{25.07} &\multicolumn{1}{c}{47.99} &\multicolumn{1}{c}{32.08} &\multicolumn{1}{c}{\bf39.69} 
\\
\cmidrule(lr){2-16}
\multicolumn{1}{c}{C2} &\multicolumn{1}{c}{43.93} &\multicolumn{1}{c}{38.64} &\multicolumn{1}{c}{40.50} &\multicolumn{1}{c}{44.62} &\multicolumn{1}{c}{33.04} &\multicolumn{1}{c}{34.69} &\multicolumn{1}{c}{35.51} &\multicolumn{1}{c}{41.44} &\multicolumn{1}{c}{26.64} &\multicolumn{1}{c}{21.65} &\multicolumn{1}{c}{36.43} &\multicolumn{1}{c}{14.20} &\multicolumn{1}{c}{44.59} &\multicolumn{1}{c}{18.91} &\multicolumn{1}{c}{28.06} 
\\
\multicolumn{1}{c}{C2 + \blice} &\multicolumn{1}{c}{\bf46.14} &\multicolumn{1}{c}{\bf47.02} &\multicolumn{1}{c}{\bf51.21} &\multicolumn{1}{c}{\bf50.17} &\multicolumn{1}{c}{\bf46.91} &\multicolumn{1}{c}{\bf41.60} &\multicolumn{1}{c}{\bf45.24} &\multicolumn{1}{c}{\bf44.59} &\multicolumn{1}{c}{\bf37.60} &\multicolumn{1}{c}{\bf33.91} &\multicolumn{1}{c}{\bf42.96} &\multicolumn{1}{c}{\bf28.80} &\multicolumn{1}{c}{\bf48.44} &\multicolumn{1}{c}{\bf33.62} &\multicolumn{1}{c}{39.24} 
\\

\toprule
\rowcolor{Gray}
\multicolumn{1}{c}{[1k] \bf Pairs - Second Half}  &\multicolumn{1}{c}{\bf CA$\to$BG}  &\multicolumn{1}{c}{\bf HE$\to$BG} &\multicolumn{1}{c}{\bf ET$\to$BG} &\multicolumn{1}{c}{\bf HU$\to$BG}
&\multicolumn{1}{c}{\bf KA$\to$BG}
&\multicolumn{1}{c}{\bf HE$\to$CA}
&\multicolumn{1}{c}{\bf ET$\to$CA}
&\multicolumn{1}{c}{\bf HU$\to$CA}
&\multicolumn{1}{c}{\bf KA$\to$CA}
&\multicolumn{1}{c}{\bf ET$\to$HE}
&\multicolumn{1}{c}{\bf HU$\to$HE}
&\multicolumn{1}{c}{\bf KA$\to$HE}
&\multicolumn{1}{c}{\bf HU$\to$ET}
&\multicolumn{1}{c}{\bf KA$\to$ET}
&\multicolumn{1}{c}{\bf KA$\to$HU}\\ 

\cmidrule(lr){2-16}
\multicolumn{1}{c}{RCSLS} &\multicolumn{1}{c}{18.53} &\multicolumn{1}{c}{14.22} &\multicolumn{1}{c}{15.01} &\multicolumn{1}{c}{18.95} &\multicolumn{1}{c}{10.10} &\multicolumn{1}{c}{12.87} &\multicolumn{1}{c}{9.34} &\multicolumn{1}{c}{19.91} &\multicolumn{1}{c}{7.17} &\multicolumn{1}{c}{5.03} &\multicolumn{1}{c}{9.73} &\multicolumn{1}{c}{2.87} &\multicolumn{1}{c}{15.81} &\multicolumn{1}{c}{6.27} &\multicolumn{1}{c}{7.59} 
\\
\multicolumn{1}{c}{RCSLS + \blice} &\multicolumn{1}{c}{33.37} &\multicolumn{1}{c}{29.14} &\multicolumn{1}{c}{31.49} &\multicolumn{1}{c}{35.14} &\multicolumn{1}{c}{26.62} &\multicolumn{1}{c}{25.33} &\multicolumn{1}{c}{23.04} &\multicolumn{1}{c}{34.93} &\multicolumn{1}{c}{16.61} &\multicolumn{1}{c}{14.78} &\multicolumn{1}{c}{21.21} &\multicolumn{1}{c}{7.34} &\multicolumn{1}{c}{32.66} &\multicolumn{1}{c}{19.76} &\multicolumn{1}{c}{18.11} 
\\
\cmidrule(lr){2-16}
\multicolumn{1}{c}{\vecmap} &\multicolumn{1}{c}{33.54} &\multicolumn{1}{c}{31.14} &\multicolumn{1}{c}{30.19} &\multicolumn{1}{c}{36.52} &\multicolumn{1}{c}{25.37} &\multicolumn{1}{c}{34.32} &\multicolumn{1}{c}{26.08} &\multicolumn{1}{c}{39.88} &\multicolumn{1}{c}{23.15} &\multicolumn{1}{c}{22.47} &\multicolumn{1}{c}{24.89} &\multicolumn{1}{c}{19.01} &\multicolumn{1}{c}{30.03} &\multicolumn{1}{c}{21.89} &\multicolumn{1}{c}{24.38} 
\\
\multicolumn{1}{c}{\vecmap + \blice} &\multicolumn{1}{c}{38.19} &\multicolumn{1}{c}{39.31} &\multicolumn{1}{c}{43.73} &\multicolumn{1}{c}{46.31} &\multicolumn{1}{c}{35.47} &\multicolumn{1}{c}{38.55} &\multicolumn{1}{c}{37.40} &\multicolumn{1}{c}{45.71} &\multicolumn{1}{c}{32.48} &\multicolumn{1}{c}{36.71} &\multicolumn{1}{c}{30.83} &\multicolumn{1}{c}{\bf33.56} &\multicolumn{1}{c}{38.51} &\multicolumn{1}{c}{32.54} &\multicolumn{1}{c}{35.42} 
\\
\cmidrule(lr){2-16}
\multicolumn{1}{c}{C1} &\multicolumn{1}{c}{38.13} &\multicolumn{1}{c}{33.67} &\multicolumn{1}{c}{33.96} &\multicolumn{1}{c}{39.86} &\multicolumn{1}{c}{29.23} &\multicolumn{1}{c}{34.84} &\multicolumn{1}{c}{29.89} &\multicolumn{1}{c}{43.89} &\multicolumn{1}{c}{24.69} &\multicolumn{1}{c}{18.78} &\multicolumn{1}{c}{33.14} &\multicolumn{1}{c}{12.21} &\multicolumn{1}{c}{38.02} &\multicolumn{1}{c}{19.53} &\multicolumn{1}{c}{26.05} 
\\
\multicolumn{1}{c}{C1 + \blice} &\multicolumn{1}{c}{44.19} &\multicolumn{1}{c}{42.19} &\multicolumn{1}{c}{46.26} &\multicolumn{1}{c}{\bf48.16} &\multicolumn{1}{c}{42.96} &\multicolumn{1}{c}{\bf42.09} &\multicolumn{1}{c}{43.43} &\multicolumn{1}{c}{49.45} &\multicolumn{1}{c}{35.72} &\multicolumn{1}{c}{36.28} &\multicolumn{1}{c}{41.74} &\multicolumn{1}{c}{23.62} &\multicolumn{1}{c}{47.37} &\multicolumn{1}{c}{33.55} &\multicolumn{1}{c}{37.78} 
\\
\cmidrule(lr){2-16}
\multicolumn{1}{c}{C2} &\multicolumn{1}{c}{40.06} &\multicolumn{1}{c}{38.37} &\multicolumn{1}{c}{38.49} &\multicolumn{1}{c}{43.03} &\multicolumn{1}{c}{34.94} &\multicolumn{1}{c}{39.59} &\multicolumn{1}{c}{39.34} &\multicolumn{1}{c}{47.14} &\multicolumn{1}{c}{27.08} &\multicolumn{1}{c}{25.20} &\multicolumn{1}{c}{36.09} &\multicolumn{1}{c}{12.94} &\multicolumn{1}{c}{44.58} &\multicolumn{1}{c}{23.79} &\multicolumn{1}{c}{27.37} 
\\
\multicolumn{1}{c}{C2 + \blice} &\multicolumn{1}{c}{\bf44.76} &\multicolumn{1}{c}{\bf43.60} &\multicolumn{1}{c}{\bf47.44} &\multicolumn{1}{c}{47.98} &\multicolumn{1}{c}{\bf45.63} &\multicolumn{1}{c}{41.97} &\multicolumn{1}{c}{\bf47.40} &\multicolumn{1}{c}{\bf50.61} &\multicolumn{1}{c}{\bf36.18} &\multicolumn{1}{c}{\bf39.25} &\multicolumn{1}{c}{\bf43.21} &\multicolumn{1}{c}{24.28} &\multicolumn{1}{c}{\bf47.65} &\multicolumn{1}{c}{\bf36.09} &\multicolumn{1}{c}{\bf38.13} 
\\

\bottomrule
\end{tabular}
}

\caption{BLI scores (P@1$\times100\%$) on the PanLex-BLI benchmark, consiting of six lower-resource languages. We apply the proposed \blice to the four CLWE baselines (compare the `baseline + \blice' results with the results from the `raw' baselines).}
\label{table:appendix-panlex}
\end{center}
\end{table*}

\section{Values of the $\lambda$ Hyper-parameter}
\label{appendix:lambda}

The hyper-parameter values of $\lambda$ are tuned on (\textsc{en}, \textsc{tr}) and (\textsc{hu}, \textsc{ka}) translation directions for XLING and PanLex-BLI, respectively; $\lambda \in \{0, 0.01, 0.02, ... , 0.98, 0.99, 1\}$. Here, we show the finally selected $\lambda$ values in our experiments, spanning the results in \S\ref{s:results} and Appendix~\ref{s:unsup}.

\begin{table*}[!t]
\begin{center}
\def\arraystretch{0.99}
\resizebox{0.9\textwidth}{!}{%
\begin{tabular}{lllllll}
\toprule 

\rowcolor{Gray}
\multicolumn{1}{c}{\bf $\lambda$ Values}  &\multicolumn{1}{c}{\bf XLING: [5k] Pairs}  &\multicolumn{1}{c}{\bf XLING: [1k] Pairs} &\multicolumn{1}{c}{\bf XLING: Unsupervised} &\multicolumn{1}{c}{\bf XLING: Zero-shot} &\multicolumn{1}{c}{\bf PanLex-BLI: [1k] Pairs}

\\ \cmidrule(lr){2-6}
\multicolumn{1}{c}{RCSLS + \blice (XLM-R$_{large}$)}  &\multicolumn{1}{c}{0.29} &\multicolumn{1}{c}{0.82} &\multicolumn{1}{c}{-} &\multicolumn{1}{c}{-} &\multicolumn{1}{c}{0.74}

\\ \cmidrule(lr){2-6}
\multicolumn{1}{c}{\vecmap + \blice (XLM-R$_{large}$)}  &\multicolumn{1}{c}{0.36} &\multicolumn{1}{c}{0.61} &\multicolumn{1}{c}{0.68} &\multicolumn{1}{c}{0.68} &\multicolumn{1}{c}{0.65}

\\ \cmidrule(lr){2-6}

\multicolumn{1}{c}{C1 + \blice (mBERT)}  &\multicolumn{1}{c}{0.18} &\multicolumn{1}{c}{0.38} &\multicolumn{1}{c}{-} &\multicolumn{1}{c}{-} &\multicolumn{1}{c}{-}  \\

\multicolumn{1}{c}{C1 + \blice (XLM-R$_{base}$)} &\multicolumn{1}{c}{0.22} &\multicolumn{1}{c}{0.40} &\multicolumn{1}{c}{-} &\multicolumn{1}{c}{-} &\multicolumn{1}{c}{-}  \\

\multicolumn{1}{c}{C1 + \blice (XLM-R$_{large}$)}  &\multicolumn{1}{c}{0.35} &\multicolumn{1}{c}{0.51} &\multicolumn{1}{c}{-} &\multicolumn{1}{c}{-} &\multicolumn{1}{c}{0.65}\\

\multicolumn{1}{c}{C1 + \blice (XLM-R$_{large}$, off-the-shelf)}  &\multicolumn{1}{c}{0.82} &\multicolumn{1}{c}{0.66} &\multicolumn{1}{c}{-} &\multicolumn{1}{c}{-} &\multicolumn{1}{c}{-}\\

\multicolumn{1}{c}{C1 + \blice (XLM-R$_{large}$, w/o Template)}  &\multicolumn{1}{c}{0.22} &\multicolumn{1}{c}{0.46} &\multicolumn{1}{c}{-} &\multicolumn{1}{c}{-} &\multicolumn{1}{c}{-}\\

\multicolumn{1}{c}{C1 + \blice (XLM-R$_{large}$, $\alpha$=$0.0$)}  &\multicolumn{1}{c}{0.09} &\multicolumn{1}{c}{0.20} &\multicolumn{1}{c}{-} &\multicolumn{1}{c}{-} &\multicolumn{1}{c}{-}\\

\multicolumn{1}{c}{C1 + \blice (XLM-R$_{large}$, $\alpha$=$0.1$)}  &\multicolumn{1}{c}{0.15} &\multicolumn{1}{c}{0.14} &\multicolumn{1}{c}{-} &\multicolumn{1}{c}{-} &\multicolumn{1}{c}{-}\\

\multicolumn{1}{c}{C1 + \blice (XLM-R$_{large}$, $\alpha$=$0.2$)}  &\multicolumn{1}{c}{0.11} &\multicolumn{1}{c}{0.21} &\multicolumn{1}{c}{-} &\multicolumn{1}{c}{-} &\multicolumn{1}{c}{-}\\

\multicolumn{1}{c}{C1 + \blice (XLM-R$_{large}$, $\alpha$=$0.3$)}  &\multicolumn{1}{c}{0.20} &\multicolumn{1}{c}{0.24} &\multicolumn{1}{c}{-} &\multicolumn{1}{c}{-} &\multicolumn{1}{c}{-}\\

\multicolumn{1}{c}{C1 + \blice (XLM-R$_{large}$, $\alpha$=$0.4$)}  &\multicolumn{1}{c}{0.21} &\multicolumn{1}{c}{0.32} &\multicolumn{1}{c}{-} &\multicolumn{1}{c}{-} &\multicolumn{1}{c}{-}\\

\multicolumn{1}{c}{C1 + \blice (XLM-R$_{large}$, $\alpha$=$0.5$)}  &\multicolumn{1}{c}{0.32} &\multicolumn{1}{c}{0.28} &\multicolumn{1}{c}{-} &\multicolumn{1}{c}{-} &\multicolumn{1}{c}{-}\\

\multicolumn{1}{c}{C1 + \blice (XLM-R$_{large}$, $\alpha$=$0.6$)}  &\multicolumn{1}{c}{0.19} &\multicolumn{1}{c}{0.29} &\multicolumn{1}{c}{-} &\multicolumn{1}{c}{-} &\multicolumn{1}{c}{-}\\

\multicolumn{1}{c}{C1 + \blice (XLM-R$_{large}$, $\alpha$=$0.7$)}  &\multicolumn{1}{c}{0.35} &\multicolumn{1}{c}{0.30} &\multicolumn{1}{c}{-} &\multicolumn{1}{c}{-} &\multicolumn{1}{c}{-}\\

\multicolumn{1}{c}{C1 + \blice (XLM-R$_{large}$, $\alpha$=$0.8$)}  &\multicolumn{1}{c}{0.37} &\multicolumn{1}{c}{0.33} &\multicolumn{1}{c}{-} &\multicolumn{1}{c}{-} &\multicolumn{1}{c}{-}\\

\multicolumn{1}{c}{C1 + \blice (XLM-R$_{large}$, $\alpha$=$0.9$)}  &\multicolumn{1}{c}{0.45} &\multicolumn{1}{c}{0.43} &\multicolumn{1}{c}{-} &\multicolumn{1}{c}{-} &\multicolumn{1}{c}{-}\\

\multicolumn{1}{c}{C1 + \blice (XLM-R$_{large}$, $\alpha$=$1.0$)}  &\multicolumn{1}{c}{0.43} &\multicolumn{1}{c}{0.51} &\multicolumn{1}{c}{-} &\multicolumn{1}{c}{-} &\multicolumn{1}{c}{-}\\

\cmidrule(lr){2-6}

\multicolumn{1}{c}{C2 + \blice (mBERT)} &\multicolumn{1}{c}{0.17} &\multicolumn{1}{c}{0.17} &\multicolumn{1}{c}{-} &\multicolumn{1}{c}{-} &\multicolumn{1}{c}{-}\\

\multicolumn{1}{c}{C2 + \blice (XLM-R$_{base}$)} &\multicolumn{1}{c}{0.15} &\multicolumn{1}{c}{0.21} &\multicolumn{1}{c}{-} &\multicolumn{1}{c}{-} &\multicolumn{1}{c}{-}\\

\multicolumn{1}{c}{C2 + \blice (XLM-R$_{large}$)}  &\multicolumn{1}{c}{0.31} &\multicolumn{1}{c}{0.50} &\multicolumn{1}{c}{-} &\multicolumn{1}{c}{-} &\multicolumn{1}{c}{0.57} \\

\multicolumn{1}{c}{C2 + \blice (XLM-R$_{large}$, off-the-shelf)}  &\multicolumn{1}{c}{0.67} &\multicolumn{1}{c}{0} &\multicolumn{1}{c}{-} &\multicolumn{1}{c}{-} &\multicolumn{1}{c}{-}\\

\multicolumn{1}{c}{C2 + \blice (XLM-R$_{large}$, w/o Template)}  &\multicolumn{1}{c}{0.25} &\multicolumn{1}{c}{0.43} &\multicolumn{1}{c}{-} &\multicolumn{1}{c}{-} &\multicolumn{1}{c}{-}\\

\multicolumn{1}{c}{C2 + \blice (XLM-R$_{large}$, $\alpha$=$0.0$)}  &\multicolumn{1}{c}{0.14} &\multicolumn{1}{c}{0.14} &\multicolumn{1}{c}{-} &\multicolumn{1}{c}{-} &\multicolumn{1}{c}{-}\\

\multicolumn{1}{c}{C2 + \blice (XLM-R$_{large}$, $\alpha$=$0.1$)}  &\multicolumn{1}{c}{0.16} &\multicolumn{1}{c}{0.15} &\multicolumn{1}{c}{-} &\multicolumn{1}{c}{-} &\multicolumn{1}{c}{-}\\

\multicolumn{1}{c}{C2 + \blice (XLM-R$_{large}$, $\alpha$=$0.2$)}  &\multicolumn{1}{c}{0.17} &\multicolumn{1}{c}{0.17} &\multicolumn{1}{c}{-} &\multicolumn{1}{c}{-} &\multicolumn{1}{c}{-}\\

\multicolumn{1}{c}{C2 + \blice (XLM-R$_{large}$, $\alpha$=$0.3$)}  &\multicolumn{1}{c}{0.20} &\multicolumn{1}{c}{0.19} &\multicolumn{1}{c}{-} &\multicolumn{1}{c}{-} &\multicolumn{1}{c}{-}\\

\multicolumn{1}{c}{C2 + \blice (XLM-R$_{large}$, $\alpha$=$0.4$)}  &\multicolumn{1}{c}{0.22} &\multicolumn{1}{c}{0.17} &\multicolumn{1}{c}{-} &\multicolumn{1}{c}{-} &\multicolumn{1}{c}{-}\\

\multicolumn{1}{c}{C2 + \blice (XLM-R$_{large}$, $\alpha$=$0.5$)}  &\multicolumn{1}{c}{0.26} &\multicolumn{1}{c}{0.26} &\multicolumn{1}{c}{-} &\multicolumn{1}{c}{-} &\multicolumn{1}{c}{-}\\

\multicolumn{1}{c}{C2 + \blice (XLM-R$_{large}$, $\alpha$=$0.6$)}  &\multicolumn{1}{c}{0.26} &\multicolumn{1}{c}{0.22} &\multicolumn{1}{c}{-} &\multicolumn{1}{c}{-} &\multicolumn{1}{c}{-}\\

\multicolumn{1}{c}{C2 + \blice (XLM-R$_{large}$, $\alpha$=$0.7$)}  &\multicolumn{1}{c}{0.31} &\multicolumn{1}{c}{0.28} &\multicolumn{1}{c}{-} &\multicolumn{1}{c}{-} &\multicolumn{1}{c}{-}\\

\multicolumn{1}{c}{C2 + \blice (XLM-R$_{large}$, $\alpha$=$0.8$)}  &\multicolumn{1}{c}{0.42} &\multicolumn{1}{c}{0.33} &\multicolumn{1}{c}{-} &\multicolumn{1}{c}{-} &\multicolumn{1}{c}{-}\\

\multicolumn{1}{c}{C2 + \blice (XLM-R$_{large}$, $\alpha$=$0.9$)}  &\multicolumn{1}{c}{0.35} &\multicolumn{1}{c}{0.43} &\multicolumn{1}{c}{-} &\multicolumn{1}{c}{-} &\multicolumn{1}{c}{-}\\

\multicolumn{1}{c}{C2 + \blice (XLM-R$_{large}$, $\alpha$=$1.0$)}  &\multicolumn{1}{c}{0.44} &\multicolumn{1}{c}{0.50} &\multicolumn{1}{c}{-} &\multicolumn{1}{c}{-} &\multicolumn{1}{c}{-}\\

\bottomrule
\end{tabular}
}

%\vspace{-1.5mm}
\caption{$\lambda$ values. The cells with `-' represent BLI setups not covered in our experiments: only unsupervised \vecmap is used for XLING unsupervised and zero-shot setups; we conduct ablation studies and investigate model variants on XLING 5k and 1k setups only.}
\label{table:lambda}
\end{center}
%\vspace{-1mm}
\end{table*}

\section{Reproducibility Checklist}
\label{app:reproducibility}
\begin{itemize}
    \item \textbf{BLI Data}: BLI datasets used in our experiments are publicly available.\footnote{\url{https://github.com/codogogo/xling-eval}} \footnote{\url{https://github.com/cambridgeltl/panlex-bli}}
    \item \textbf{Static Word Embeddings}: We adopt the standard monolingual word embeddings for deriving CLWEs, used in a body of prior work on BLI. In fact, the XLING benchmark already provides a set of preprocessed fastText WEs trained on Wikipedia, which is then our starting point.\footnote{\url{https://fasttext.cc/docs/en/pretrained-vectors.html}} Panlex-BLI does not provide processed WEs, so we follow the original instructions from the authors and adopt fastText WEs trained on Common Crawl $+$ Wikipedia.\footnote{\url{https://fasttext.cc/docs/en/crawl-vectors.html}} The WEs are trimmed to the most frequent $200k$ words for each language. The same WEs are used for all CLWE baselines.
    \item \textbf{Pretrained LMs}: We derive CEs by fine-tuning pretrained LMs includings the mBERT variant `bert-base-multilingual-uncased', and the XLM-R variants `xlm-roberta-base' and `xlm-roberta-large', all publicly available from the \href{http://huggingface.co}{huggingface.co} model hub.    
    \item \textbf{Parameter Counts}: The number of parameters are $167,356,416$ for mBERT, $278,043,648$ for XLM-R$_{base}$, and $559,890,432$ for XLM-R$_{large}$.
    \item \textbf{Source Code}: We release our code at \url{https://github.com/cambridgeltl/BLICEr}.
    \item \textbf{Computing Infrastructure}: We run our code on \href{https://www.hpc.cam.ac.uk/systems/wilkes-3}{Wilkes3}, a cluster with $80$ nodes. Each node has $4$ $\times$ Nvidia 80GB A100 GPUs and $128$ $\times$ CPU cores. All our experiments require only one node with $1\times$ GPU and $32\times$ CPU cores. 
    \item \textbf{Software}: Slurm $20.11.8$, Python $3.9.7$, PyTorch $1.10.1$, Transformers $4.15.0$, and Sentence-Transformers $2.1.0$.
    \item \textbf{Runtime (Wall Time)}: The \blice training (XLM-R$_{large}$, C2 as the CLWE backbone) on a language pair typically costs $10$ minutes in both supervised and semi-supervised BLI setups. It takes $3$ minutes for one BLI evaluation run. 
    \item \textbf{Robustness}: We found that the improvements of \blice are robust over all language pairs with different random seeds, and thus use a fixed random seed $33$ over all experiments.   
\end{itemize}

\end{document}